\documentclass[letterpaper,journal]{IEEEtran}
\usepackage{amsmath,amsfonts,amssymb}
\usepackage{array,booktabs,multirow,tabularx,placeins}
\usepackage[caption=false,font=footnotesize]{subfig}
\usepackage{textcomp,stfloats,url,graphicx,cite}
\usepackage[hidelinks]{hyperref}
\usepackage{capt-of}
\newlength{\ablationdatasetwidth}
\newlength{\ablationmethodwidth}
\makeatletter
\newenvironment{teaserfigure}{%
    \par\noindent\begin{minipage}[b]{\textwidth}%
    \def\@captype{figure}\centering
}{\end{minipage}\par}
\makeatother
\begin{document}
\title{OctreeSplat: Hierarchical Octree-Guided\\Feed-Forward Gaussian Splatting}

\author{Yibin Zhao\textsuperscript{*}, Yihan Pan\textsuperscript{*}, Wenli Yang, Jun Nan, and Jianjun Yi\textsuperscript{\ensuremath{\dagger}}%
\thanks{\textsuperscript{*}: equal contribution.}%
\thanks{\textsuperscript{\ensuremath{\dagger}}: Corresponding author.}%
\thanks{Yibin Zhao, Wenli Yang, Jun Nan, and Jianjun Yi: East China University of Science and Technology, Shanghai, China (e-mail: \{Y20230063,Y30260694,Y20240039\}@mail.ecust.edu.cn; jjyi@ecust.edu.cn).}%
\thanks{Yihan Pan: Shanghai Open University, Shanghai, China (e-mail: pyh@sou.edu.cn).}}

\IEEEaftertitletext{%
\begin{teaserfigure}
    \centering
    \includegraphics[width=\linewidth]{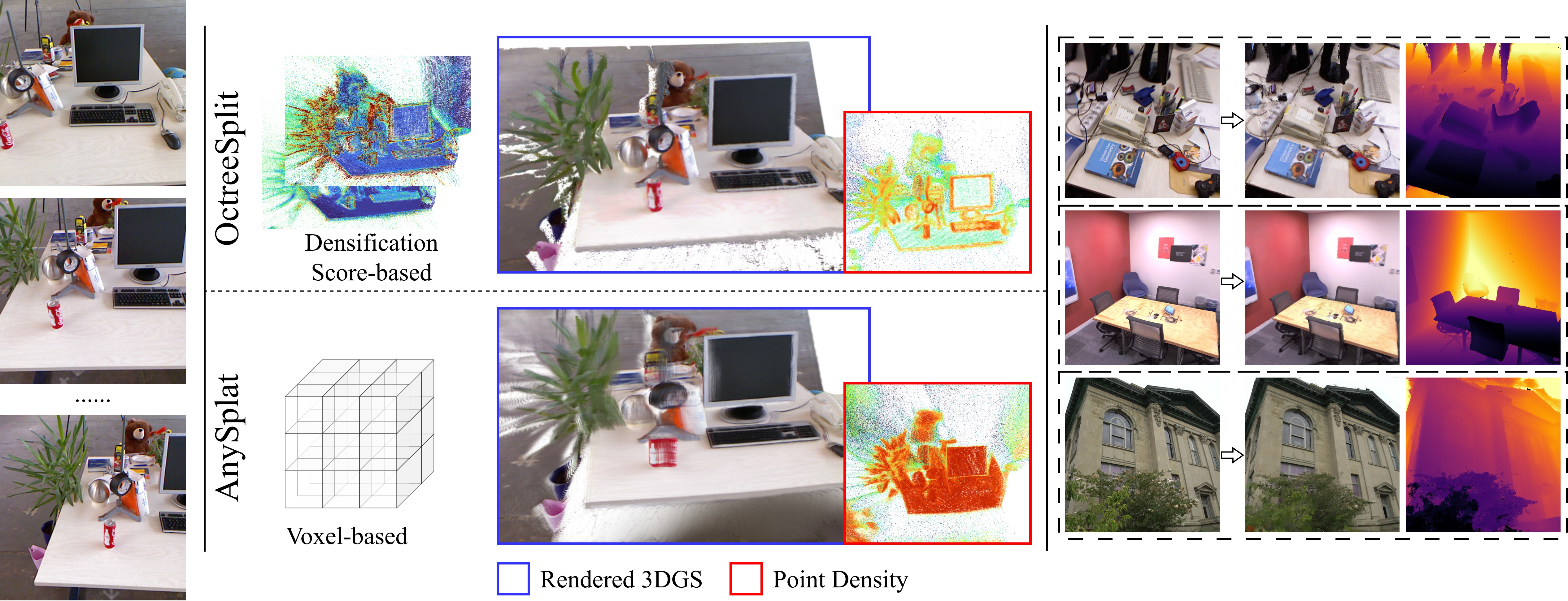}
    \caption{Given a variable number of images and optional camera parameters, \textbf{OctreeSplat} reconstructs a geometrically accurate 3DGS scene through feed-forward inference. Predicted densification scores guide the spatial allocation of Gaussian primitives. Compared with pixel-aligned\cite{depthanything3} and voxel-aligned\cite{jiang2025anysplat} methods, OctreeSplat improves RGB synthesis, depth synthesis, and surface reconstruction while using fewer primitives. Optional camera-conditioned refinement is performed through test-time optimization.}
    \label{fig:toutu}
\end{teaserfigure}
    \vspace{0.5em}
}
\maketitle
\begin{abstract}
3D Gaussian Splatting (3DGS) has shown considerable potential for novel view synthesis (NVS) and 3D reconstruction. Feed-forward methods avoid costly per-scene optimization, but fixed image or spatial resolutions limit their ability to allocate primitives according to local complexity. We present OctreeSplat, a hierarchical octree-guided feed-forward Gaussian splatting. OctreeSplat extracts multi-view features and geometric priors from a frozen visual foundation model (VFM), constructs a hierarchical octree, and predicts densification scores. The resulting representation allocates Gaussian primitives at spatially varying densities according to local complexity. Stochastic solids volume rendering improves depth rendering, strengthens geometric supervision, and supports surface reconstruction. When camera parameters are available, an optional test-time optimization (TTO) module improves agreement between predicted and ground-truth parameters. Training updates only 76 million additional parameters and requires 35 GPU-hours. Experiments on Replica, TUM, Tanks and Temples, and DTU demonstrate competitive RGB view synthesis, depth synthesis, and surface reconstruction. OctreeSplat reduces the number of Gaussian primitives while maintaining high reconstruction quality.
\end{abstract}


\section{Introduction}
Image-based 3D reconstruction has traditionally relied on structure-from-motion (SfM) pipelines, with iterative techniques represented by bundle adjustment (BA)\cite{BA} for multi-view pose estimation and sparse reconstruction. Common scene representations include point clouds, voxels, and signed distance fields. More recently, 3D Gaussian Splatting (3DGS)\cite{kerbl3DGS} has combined photorealistic novel view synthesis with high rendering efficiency. Given images and camera parameters, 3DGS iteratively renders the scene and optimizes the parameters of Gaussian primitives. Adaptive density control (ADC) periodically adds and removes primitives during optimization, which typically takes tens of minutes to an hour.

Visual foundation models (VFMs)\cite{wang2025vggt,wang2025pi3,wang2026vggtomega} learn strong priors from large-scale datasets and directly regress 3D information from image collections. Outputs include depth maps, 3D point maps, and camera parameters. These models reduce accumulated errors and iterative computation in conventional reconstruction pipelines while maintaining strong generalization. Feed-forward 3DGS methods\cite{ye2025yonosplatneedmodelfeedforward,jiang2025anysplat,depthanything3} similarly predict Gaussian scenes in a single feed-forward pass. However, commonly used pixel-aligned representations\cite{chen2024mvsplat,ye2025yonosplatneedmodelfeedforward} can introduce overlapping primitives and floating artifacts.
Voxel-aligned methods\cite{wang2025volsplat,jiang2025anysplat} and methods that regress primitives from fixed token sets\cite{li2026querysplat,li2026tokensplat} produce more compact representations. Nevertheless, their primitive allocation does not explicitly adapt to local scene complexity, limiting density flexibility and geometric accuracy.

To address the above problems, we propose \textbf{OctreeSplat}, a hierarchical octree-guided feed-forward Gaussian splatting. Given uncalibrated images and optional camera parameters, OctreeSplat predicts a geometrically accurate 3DGS scene with spatially adaptive primitive allocation. Depth maps and camera parameters predicted by a VFM backbone support voxel aggregation at multiple resolutions. A dense prediction transformer (DPT) upsamples the backbone's patch tokens to construct aligned multi-resolution voxel features.
A sparse-convolutional U-Net (SPUNet) processes these features and predicts Gaussian parameters at each level, together with densification scores between adjacent levels. During inference, subdivision starts from the coarsest voxels and selects a fixed fraction for refinement according to the predicted scores. Gaussian primitives associated with the selected octree nodes form the final octree-aligned scene.

We initialize the backbone with pretrained VFM weights and introduce an optional test-time optimization (TTO) module. When camera parameters are supplied, TTO improves alignment between the predicted parameters and the ground-truth.
Stochastic-solid volume rendering provides more accurate geometric supervision by representing Gaussian primitives as equivalent stochastic solids and volume rendering along viewing rays. This formulation improves depth rendering and geometric accuracy while supporting surface reconstruction. The backbone remains frozen during training. Only the feature-upsampling DPT, SPUNet, densification-score heads, and Gaussian prediction heads are optimized, totaling 76 million parameters. This design allows OctreeSplat to be trained in 35 GPU-hours on RTX PRO 6000.

Our main contributions are summarized as follows:
\begin{itemize}
    \item We introduce OctreeSplat, a hierarchical octree-guided method that reconstructs geometrically accurate Gaussian scenes from uncalibrated images and optional camera parameters.
    \item Predicted densification scores guide adaptive octree subdivision and allocate Gaussian primitives according to local scene complexity.
    \item A frozen VFM backbone limits training to 76 million additional parameters and 35 GPU-hours. Optional TTO aligns camera predictions with ground-truth without updating backbone weights.
    \item Extensive experiments demonstrate competitive RGB view synthesis and geometric reconstruction across multiple datasets and settings.
\end{itemize}

\section{Related Work}
\subsection{3D Scene Representations}
Conventional reconstruction methods represent scenes using point clouds\cite{colmapsfm}, voxels\cite{seitz1999photorealistic}, signed distance fields (SDFs)\cite{curless1996volumetric}, or triangle meshes\cite{labatut2009robust}. Neural radiance fields (NeRFs)\cite{mildenhall2020nerf} instead use multilayer perceptrons (MLPs) to represent scenes as continuous implicit radiance fields, which map 3D positions and viewing directions to volume densities and colors. Differentiable image rendering optimizes radiance fields' parameters to produce photorealistic NVS.
Subsequent work has improved reconstruction speed\cite{mueller2022instantngp,yu2022plenoxels}, rendering quality\cite{barron2022mipnerf360,barron2021mipnerf}, and surface reconstruction\cite{wang2021neus,neus2}. Despite NeRF's reconstruction accuracy, implicit representations require repeated field inferences along each ray during volume rendering. This computational cost limits rendering efficiency and transferability.

3DGS\cite{kerbl3DGS} addresses this bottleneck by explicitly storing scenes as 3D Gaussian primitives. These primitives are initialized from sparse point clouds, typically obtained through SfM\cite{colmapsfm}, and optimized by iteratively rendering the input images. ADC splits, clones, and prunes primitives using photometric gradients\cite{kerbl3DGS} or alternative criteria\cite{kim2024color,lyu2025resgs}. The optimization process typically takes tens of minutes to an hour.
Later methods accelerate optimization\cite{chen2025dashgaussian,mallick2024taming3dgs,meuleman2025fly} by controlling the primitive count. However, they still require precomputed camera poses and sparse point clouds, and optimization generally takes at least several minutes. ADC strategies have also received sustained attention\cite{ye2024absgs,ren2026fastgs}.

Vanilla 3DGS primarily optimizes RGB rendering and lacks explicit surface supervision. Subsequent studies improve geometric accuracy through surface-alignment regularization\cite{guedon2023sugar}. Depth-distortion, normal-consistency, and multi-view-consistency constraints further reduce geometric ambiguities arising from photometric supervision\cite{Huang2DGS2024,chen2024pgsr}. GOF\cite{Yu2024GOF} constructs an opacity field from Gaussian primitives and extracts its isosurface for surface reconstruction. GGGS\cite{zhang2026gggs} and GW\cite{gw} further improve surface reconstruction by representing Gaussian primitives as equivalent stochastic solids.

\subsection{Feed-Forward Reconstruction}
Another line of research predicts Gaussian scenes through a single feed-forward based on neural-network.
Feed-forward methods such as pixelSplat\cite{charatan2024pixelsplat} take images and camera parameters as input. They aggregate multi-view features, estimate per-pixel depth, back-project pixels into world coordinates, and regress Gaussian parameters. Multi-view cost volumes\cite{chen2024mvsplat} and pretrained depth priors\cite{xu2024depthsplat} improve reconstruction quality in challenging scenes. VolSplat\cite{wang2025volsplat} replaces pixel alignment with voxel alignment, reducing the dependence of primitives on individual input views.

VFMs extend reconstruction to uncalibrated images without requiring camera parameters as input. DUSt3R\cite{dust3r_cvpr24} formulates image-pair reconstruction as 3D point-map regression, allowing dense geometry prediction without camera parameters. MUSt3R\cite{cabon2025must3r} extends DUSt3R to multiple views and predicts geometry directly in a common coordinate system.
VGGT\cite{wang2025vggt} alternates single-view and multi-view self-attention to aggregate information across a variable number of views. A shared backbone and separate prediction heads jointly estimate camera parameters, depth maps, 3D point maps, and point tracks. Subsequent work improves robustness\cite{wang2025pi3}, long-sequence processing\cite{liu2025vggtx}, and model scalability\cite{wang2026vggtomega}.

VFMs have also advanced feed-forward 3DGS.Some approaches build directly upon VFM backbones by adding prediction heads for pixel-aligned Gaussian attributes\cite{depthanything3,smart2024splatt3r}. As more views are added, overlapping observations and the one-to-one pixel-to-primitive correspondence introduce redundant, overlapping primitives that can blur the rendered images.
AnySplat\cite{jiang2025anysplat} uses pretrained VGGT geometry for pseudo-geometric supervision and differentiable voxelization for multi-view aggregation, reducing redundancy and overlap artifacts. Other methods\cite{li2026querysplat,li2026tokensplat} regress Gaussian primitives directly from backbone tokens, removing the one-to-one correspondence with pixels. Nevertheless, increasing the number of input views can still introduce blur, and geometric accuracy remains limited. Some methods have explored Gaussian self-splitting during the feed-forward process\cite{kim2026f4splat}. However, they remain focused on the pixel coordinate system, which leads to the overlapping of primitives.

\section{Method}
\begin{figure*}[!tbp]
    \centering
    \includegraphics[width=\linewidth]{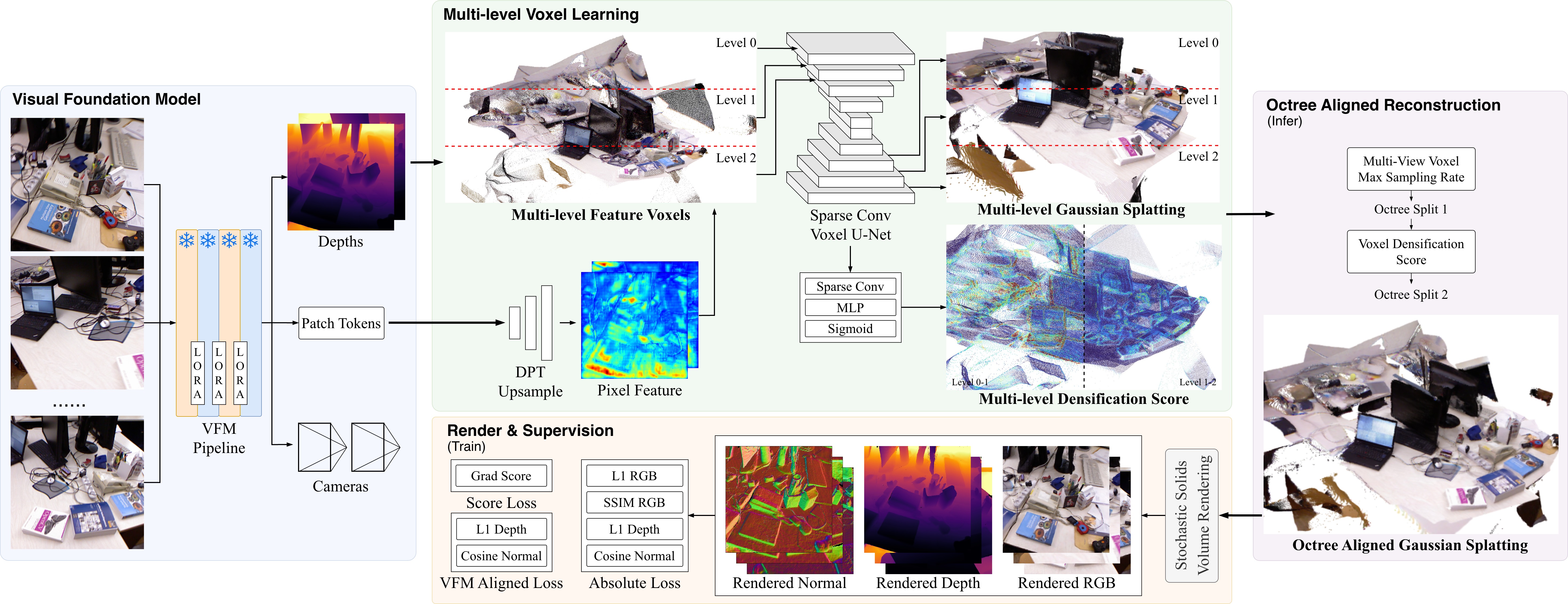}
    \caption{Overview of the \textbf{OctreeSplat} pipeline. A frozen visual foundation model extracts multi-view features and predicts geometry. DPT and SPUNet construct multi-resolution voxel features and predict Gaussian parameters and densification scores, which guide adaptive octree subdivision to construct an octree-aligned Gaussian representation that adapts to local scene complexity. Stochastic-solid volume rendering provides RGB and geometric supervision. }
    \label{fig:pipeline}
\end{figure*}

OctreeSplat reconstructs a geometrically accurate 3DGS scene from $N$ uncalibrated images and optional camera parameters, as shown in \autoref{fig:pipeline}. The network predicts Gaussian primitives at multiple octree levels and uses densification scores to guide subdivision. The hierarchical representation improves geometric accuracy and NVS quality while controlling the total number of Gaussian primitives.

\subsection{Preliminaries}

Given uncalibrated images $\mathcal{I}=\{I_i\}_{i=1}^{N}$, a VFM uses a large transformer $\mathcal{T}_{\theta}$ to aggregate multi-view features. Separate prediction heads $\mathcal{H}$ estimate depth maps $\hat{\mathbf{D}}$, camera parameters $\hat{\mathbf{C}}$, and other 3D attributes, as shown in \autoref{eq:pre_vfm_formulation}.
\begin{equation}
\begin{aligned}
\label{eq:pre_vfm_formulation}
(\hat{\mathbf{D}}, \hat{\mathbf{C}})
= \mathcal{H}\left(\mathcal{T}_{\theta}(\mathcal{I})\right)
\end{aligned}
\end{equation}

For NVS, VFM-based 3DGS methods further predict a Gaussian scene in a single forward pass, represented as $\mathcal{G}=\{(\boldsymbol{\mu}_i,\mathbf{R}_i,
\mathbf{s}_i,o_i,\mathbf{h}_i)\}_{i=1}^{M}$. Each primitive has a position $\boldsymbol{\mu}_i$, rotation $\mathbf{R}_i$, scale $\mathbf{s}_i$, opacity $o_i$, and spherical-harmonic color coefficients $\mathbf{h}_i$. Gaussian splatting and $\alpha$-blending produce novel views as shown in \autoref{eq:pre_gsrender}, where $\boldsymbol{\mu}'_i$ and
$\boldsymbol{\Sigma}'_i$ denote the projected pixel coordinates and covariance matrix, respectively.
\begin{equation}
\label{eq:pre_gsrender}
\begin{aligned}
\hat{\mathbf{C}}(\mathbf{p})
&=\sum_{i=1}^{M}\left(\prod_{j=1}^{i-1}(1-\alpha_j)\right)\alpha_i\mathbf{c}_i,\\
\alpha_i&=o_i\exp\!\left(-\tfrac{1}{2}(\mathbf{p}-\boldsymbol{\mu}'_i)^\top
\boldsymbol{\Sigma}'_i{}^{-1}(\mathbf{p}-\boldsymbol{\mu}'_i)\right)
\end{aligned}
\end{equation}

A single forward pass cannot readily reproduce the iterative splitting and cloning of the original 3DGS\cite{kerbl3DGS}. Consequently, many feed-forward methods use pixel-aligned representations with fixed primitive counts. AnySplat\cite{jiang2025anysplat} introduces differentiable voxelization for voxel-aligned prediction, allowing the primitive count to be controlled through voxel size. However, assigning only one Gaussian to each voxel can reduce representation accuracy\cite{kim2026f4splat}. Methods without explicit pixel or voxel alignment\cite{li2026tokensplat,li2026querysplat} remove this correspondence constraint, but do not explicitly adapt primitive density to local scene complexity. 

\subsection{Visual Foundation Model Backbone}
OctreeSplat's VFM backbone follows the DA3 architecture\cite{depthanything3}. A pretrained DINOv2 encoder\cite{oquab2023dinov2} extracts patch tokens $\mathbf{t}_{\mathrm{patch},i}$ from each input image $I_i$. These tokens are concatenated with learnable camera tokens $\mathbf{t}_{\mathrm{cam}}$ and register tokens $\mathbf{t}_{\mathrm{reg}}$, then processed by 40 attention layers. The first 12 layers apply single-view self-attention; subsequent layers alternate single-view self-attention with global multi-view self-attention. A DPT decoder\cite{DPT} upsamples the resulting patch tokens into pixel-aligned depth maps. An MLP decodes the camera tokens into camera parameters, including poses and fields of view.

When camera parameters are available, lightweight low-rank adaptation (LoRA) modules support TTO to align predictions with ground-truth. LoRA matrices are inserted into the alternating attention layers, starting from layer 13, and applied to the query, key, and value projections. Within each attention layer, these projections share a down-projection matrix $A$ and use separate up-projection matrices $B_Q$, $B_K$, and $B_V$, as shown in \autoref{eq:lora_projection}.
\begin{equation}
\label{eq:lora_projection}
y=Wx+b+\frac{\alpha}{r}BAx
\end{equation}
Here, $x\in\mathbb{R}^{d}$ and $y\in\mathbb{R}^{d}$ are the input and output matrixs. The original projection parameters are $W\in\mathbb{R}^{d\times d}$ and $b\in\mathbb{R}^{d}$. The low-rank dimension is $r=4$ by default, and the scaling factor $\alpha/r$ is set to 4.

The LoRA modules are activated when camera parameters corresponding to the input images are supplied. A TTO loss compares the backbone's camera predictions with ground-truth. Back-propagation updates only the LoRA parameters for a fixed number of iterations. VGGT-style VFMs typically anchor the world coordinate system to the first frame, while image-based reconstruction retains an inherent scale ambiguity. We therefore align predicted and reference poses before computing the TTO loss. All poses are expressed relative to the first frame, and scale is normalized using the mean distance of the cameras.

After alignment, we compute separate rotation and translation losses. The rotation loss uses the relative rotation between predicted and reference orientations, while the translation loss applies SmoothL1 to the aligned translation vectors, as shown in \autoref{eq:tto_pose_losses}.
\begin{equation}
\label{eq:tto_pose_losses}
\begin{aligned}
\mathcal{L}_{R}&=\frac{1}{N}\sum_{i=1}^{N}\frac{3-\operatorname{tr}\!\left[\left(\widetilde{\mathbf{R}}_i^{\mathrm{pred}}\right)^{\top}\widetilde{\mathbf{R}}_i^{\mathrm{gt}}\right]}{2},\\
\mathcal{L}_{t}&=\frac{1}{N}\sum_{i=1}^{N}\operatorname{SmoothL1}\!\left(\widetilde{\mathbf{t}}_i^{\mathrm{pred}},\widetilde{\mathbf{t}}_i^{\mathrm{gt}}\right)
\end{aligned}
\end{equation}
Here, $N$ is the number of input frames. The matrices $\widetilde{\mathbf{R}}_i^{\mathrm{pred}},\widetilde{\mathbf{R}}_i^{\mathrm{gt}}\in\mathrm{SO}(3)$ denote the aligned predicted and ground-truth rotations for frame $i$. The vectors $\widetilde{\mathbf{t}}_i^{\mathrm{pred}},\widetilde{\mathbf{t}}_i^{\mathrm{gt}}\in\mathbb{R}^{3}$ denote the corresponding translations after scale alignment.
We additionally supervise camera intrinsics. For frame $i$, the intrinsic-parameter vector is $k_i=(f_{x,i},f_{y,i},c_{x,i},c_{y,i})$. An L1 loss compares the predicted and ground-truth intrinsic parameters:
\begin{equation}
\label{eq:tto_intrinsics_loss}
\mathcal{L}_{K}=\frac{1}{N}\sum_{i=1}^{N}\left\|k_i^{\mathrm{pred}}-k_i^{\mathrm{gt}}\right\|_1
\end{equation}
The complete TTO objective is
$\mathcal{L}_{\mathrm{TTO}}=\lambda_{p}\left(\mathcal{L}_{R}+\beta_t\mathcal{L}_{t}\right)+\lambda_{k}\mathcal{L}_{K}$.

\begin{figure}[!t]
    \centering
    \includegraphics[width=\linewidth]{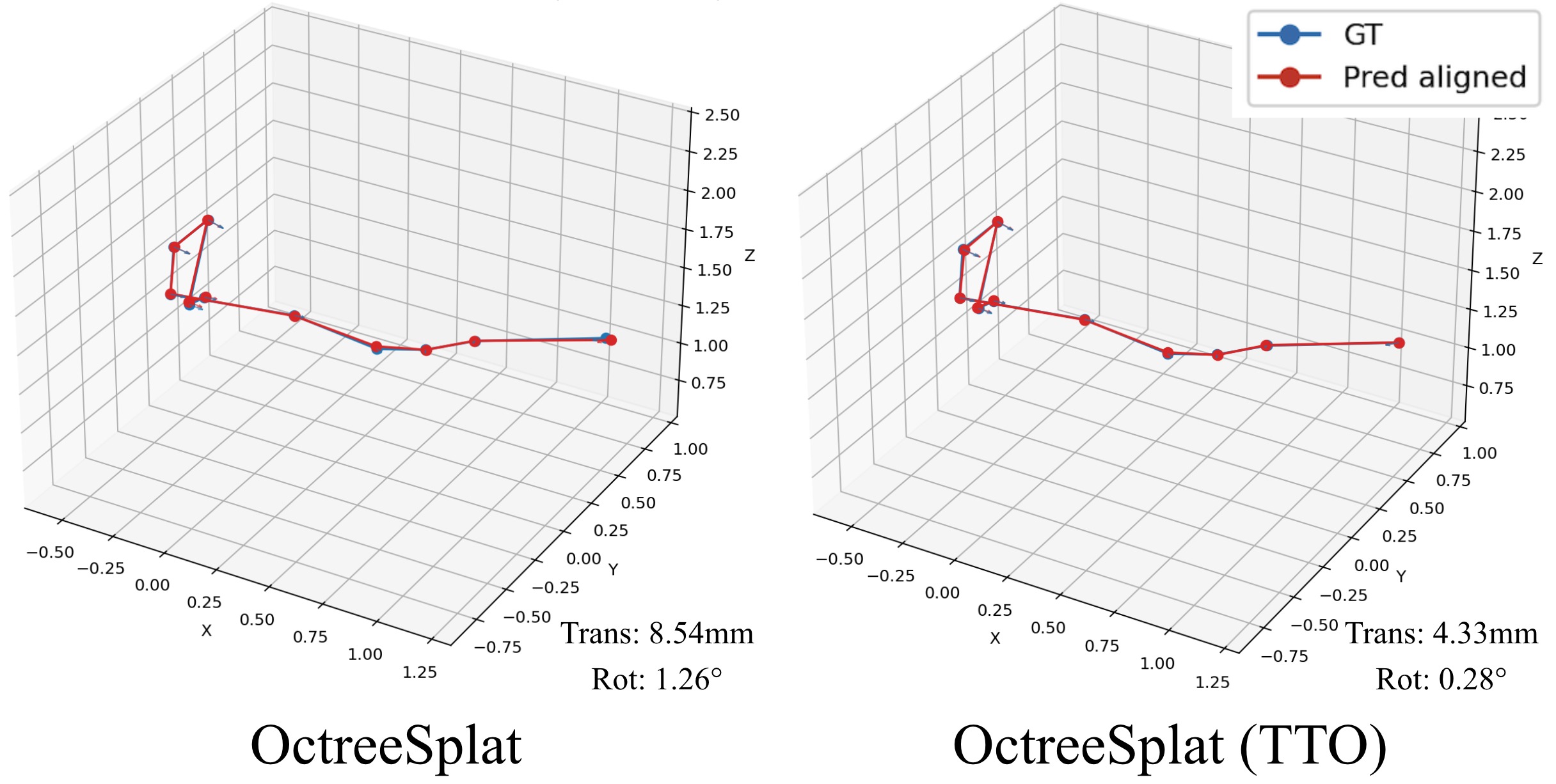}
    \caption{Improvement in pose estimation accuracy with TTO.}
    \label{fig:tto_effort}
\end{figure}

As shown in \autoref{fig:tto_effort}, TTO improves the accuracy of camera parameters predicted by the VFM backbone.

\subsection{Hierarchical Voxel Learning}

Given the VFM's depth maps $\hat{\mathbf{D}}$, patch tokens $\{\mathbf{t}_{\mathrm{patch},i}\}_{i=1}^{N}$, and camera parameters $\hat{\mathbf{C}}$, we construct hierarchical voxel features. The voxel sets $\{\mathcal{V}^{(\ell)}\}_{\ell=1}^{L}$ support prediction of Gaussian scenes $\{\mathcal{G}^{(\ell)}\}_{\ell=1}^{L}$ at the corresponding levels.

DINOv2 patch tokens $\mathbf{t}_{\mathrm{patch},i}$ have a spatial resolution downsampled by a factor of 14. To obtain dense pixel features, we introduce a feature DPT $\mathcal{D}_{\mathrm{feat}}$, that upsamples these tokens into pixel-aligned feature maps, as shown in \autoref{eq:dense_features}.
\begin{equation}
\label{eq:dense_features}
\mathbf{F}_i=\mathcal{D}_{\mathrm{feat}}(\mathbf{t}_{\mathrm{patch},i})
\end{equation}
The corresponding depth map $\hat{\mathbf{D}}_i$ and camera parameters $\hat{\mathbf{C}}_i$ are used to back-project pixels into a global 3D point cloud $\mathbf{X}_i$. Pairing each 3D point with its pixel feature produces the
global point cloud feature $\mathcal{P}=\left\{\mathbf{X}_i,\mathbf{F}_i\right\}$.

We aggregate $\mathcal{P}$ into global voxel features to obtain a compact representation. $L$ voxel levels are constructed ($L=3$), with voxel size halved at each successive level. All grids share a common origin, establishing the octree hierarchy.
For each voxel, we aggregate pixel features and 3D positions using depth-confidence weighted averaging over the included pixels $\mathcal{S}_{v}^{(\ell)}$. As shown in \autoref{eq:voxel_aggregation}, $c_i(\mathbf{p})$ denotes the confidence of the depth estimate. The aggregated 3D position replaces the voxel center to represent more accurate scene geometry.
\begin{equation}
\label{eq:voxel_aggregation}
\left\{\mathbf{f}_{v}^{(\ell)},\mathbf{x}_{v}^{(\ell)}\right\}
=\frac{
\displaystyle\sum_{(i,\mathbf{p})\in\mathcal{S}_{v}^{(\ell)}}c_i(\mathbf{p})
\left\{\mathbf{F}_i(\mathbf{p}),\mathbf{X}_i(\mathbf{p})\right\}
}{\displaystyle\sum_{(i,\mathbf{p})\in\mathcal{S}_{v}^{(\ell)}}c_i(\mathbf{p})}
\end{equation}

A sparse 3D convolutional U-Net further aggregates and enhances the scene features using SpConv\cite{spconv2022}. The first $L$ encoder stages correspond to the input voxel levels. At coarser stages, we inject the aggregated features from the matching voxel levels alongside features downsampled from finer levels. The final $L$ decoder stages predict Gaussian scenes $\{\mathcal{G}^{(\ell)}\}_{\ell=1}^{L}$ whose voxel sizes match the corresponding input levels.

The network also predicts densification scores at the $L-1$ decoder levels preceding the finest level. At each such level, a separate sparse convolution, MLP, and sigmoid function produce a score $d_v^{(\ell)}\in[0,1]$ for each voxel $v\in\mathcal{V}^{(\ell)}$.

\subsection{Octree-Aligned Reconstruction}
\begin{figure}[!h]
    \centering
    \includegraphics[width=\linewidth]{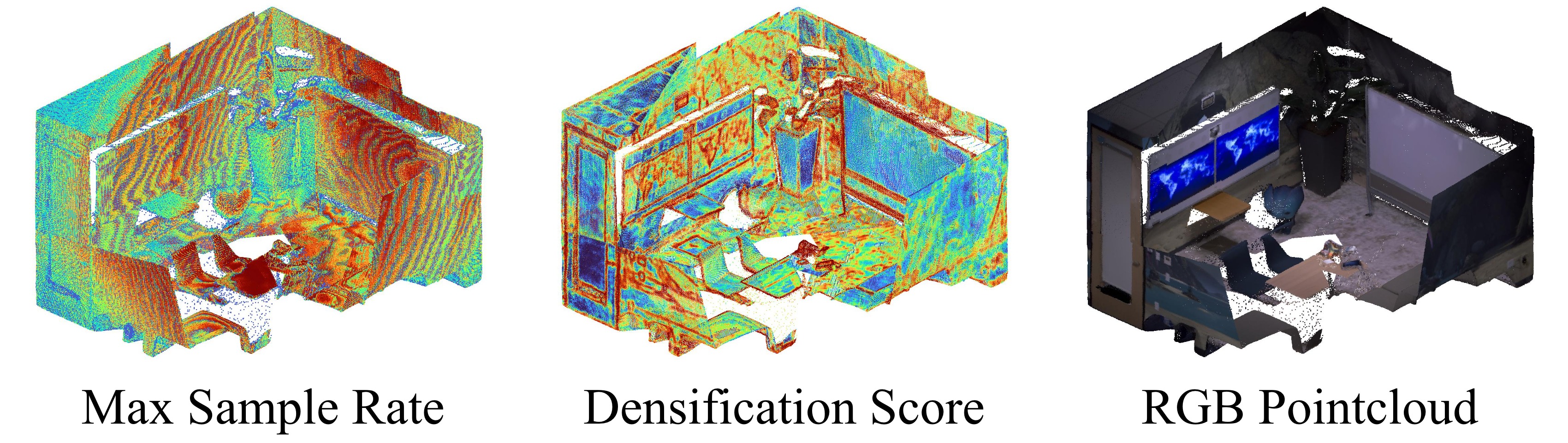}
    \caption{Comparison of the maximum multi-view sampling rate and predicted densification scores.}
    \label{fig:effort_of_score}
\end{figure}
The predicted Gaussian scenes $\{\mathcal{G}^{(\ell)}\}_{\ell=1}^{L}$ and densification scores $d_v^{(\ell)}\in[0,1]$ guide octree construction and primitive allocation. Reconstruction starts from the coarsest Gaussian scene. Previous work\cite{svraster} measures voxel complexity using the maximum multi-view sampling rate, defined as the largest number of pixels covered by a voxel across views.
However, this sampling rate reflects projected coverage rather than local detail. Nearby regions with simple textures can therefore receive high sampling rates, which may not indicate a need for Gaussian densification. We instead predict densification scores that are higher near edges and complex textures, as shown in \autoref{fig:effort_of_score}. These scores guide subdivision toward regions requiring finer representations.

At each of the $L-1$ non-finest levels, we rank voxels by their predicted scores and subdivide a fixed fraction into eight children. The corresponding finer voxels are retrieved from the next level. Children absent from that level are skipped, preserving spatial sparsity and reducing computational cost.
Repeating this procedure (2 subdivision rounds by default) constructs a sparse octree that allocates finer voxels to more complex regions. Each selected octree node is associated with a previously predicted Gaussian primitive at this level. The final scene is thus assembled from a single feed-forward prediction.

\subsection{Rendering and Training}
\subsubsection{Stochastic-Solid Volume Rendering}
\begin{figure}[!h]
    \centering
    \includegraphics[width=\linewidth]{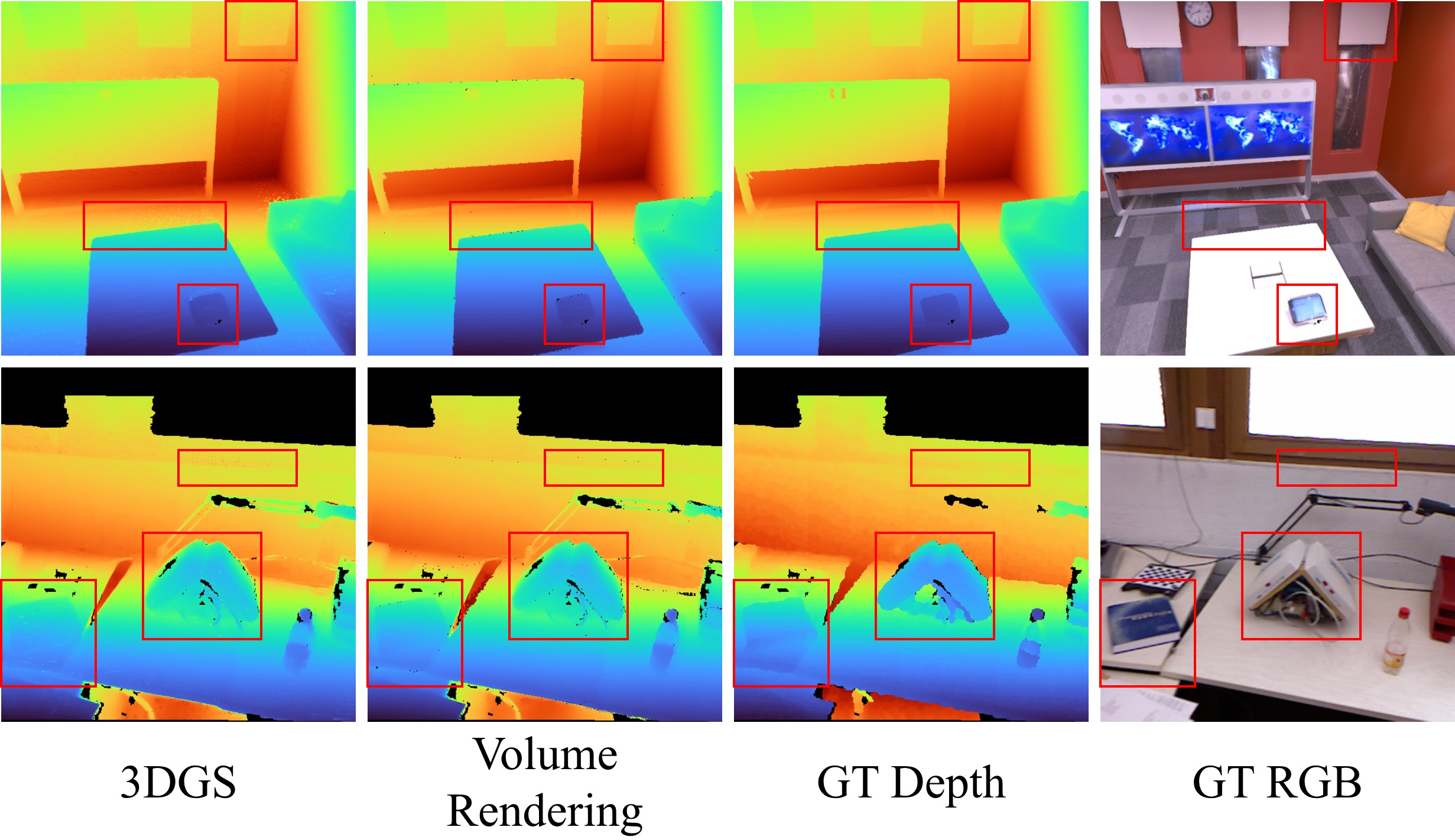}
    \caption{Depth maps rendered using standard 3DGS and stochastic-solid volume rendering for the same Gaussian scene.}
    \label{fig:effort_gggs}
\end{figure}

Vanilla 3DGS projects Gaussian primitives onto the image plane and uses $\alpha$-blending for photorealistic RGB synthesis. However, this formulation has limited accuracy for depth synthesis. We adopt stochastic-solid volume rendering\cite{2023Objectsasvolumes}, representing Gaussian primitives as equivalent spatial stochastic solids and integrating along rays. For the same Gaussian scene, this renderer improves depth accuracy and suppresses floating artifacts and erroneous depths, as shown in \autoref{fig:effort_gggs}.

Following the equivalence established in prior work\cite{zhang2026gggs}, each Gaussian primitive is replaced by an equivalent stochastic solid to preserve its RGB contribution. The associated vacancy $\mathbf{v}$ is computed as shown in \autoref{eq:vacancy}, where $G(t^{*})$ is the maximum Gaussian value along the ray.
\begin{equation}
\label{eq:vacancy}
\mathbf{v}=\sqrt{1-G(t^{*})}
\end{equation}

Both RGB images and depth maps are generated through volume rendering. The RGB rendering formulation is shwon in \autoref{eq:volume_rendering}.
\begin{equation}
\label{eq:volume_rendering}
\begin{aligned}
\hat{\mathbf{C}}(\mathbf{p})
&=\int_{t_n}^{t_f}p(t)\,\mathbf{c}(\mathbf{x}(t),\omega)\,\mathrm{d}t
=\mathbf{c}\left(1-\mathbf{v}(t)^2\right),\\
p(t)&=T(t)\,\sigma(\mathbf{x}(t),\omega),\\
T(t)&=\exp\left(-\int_{t_n}^{t}\sigma(\mathbf{x}(s),\omega)\,\mathrm{d}s\right)
\end{aligned}
\end{equation}

Depth rendering uses a fixed number of bisection iterations to locate the median depth, where accumulated transmittance $T(t)$ reaches 0.5. RGB rendering remains differentiable throughout. For depth back-propagation, we use the gradient at the median depth, with $\frac{\partial T}{\partial t}=-T(t)\sum_i\sigma_i(t)$.

\subsubsection{Training Strategy}
OctreeSplat's training pipeline is differentiable and trained end-to-end. We use pretrained DA3-GIANT weights\cite{depthanything3} for the VFM backbone. Excluding the camera-parameter encoder and Gaussian head, the backbone contains 1.190 billion parameters. OctreeSplat adds 76 million parameters: 36 million in the feature DPT and 40 million in the sparse U-Net and decoding MLPs. The backbone remains fully frozen, and only the additional parameters are optimized, substantially reducing training time.

\paragraph{Training datasets}

We train on a mixture of Infinigen\cite{infinigen2023infinite}, ScanNet++\cite{yeshwanth2023scannet++}, ARKitScenes\cite{baruch2021arkitscenes}, and DL3DV\cite{ling2024dl3dv}. Infinigen provides synthetic RGB-D data, ScanNet++ and ARKitScenes provide real-captured RGB-D data, and DL3DV provides real-captured RGB data. The combined dataset contains 6,810,309 images from 12,419 scenes, covering indoor and outdoor environments as well as synthetic and real-captured images.

\paragraph{Implementation details}
We train the additional parameters for 100,000 iterations using AdamW. The learning rate starts at $3\times10^{-6}$ and decays to $10^{-8}$ with cosine annealing. We use $L=3$ voxel levels, with a coarsest voxel size of 8 mm. Training images have a resolution of $244\times448$. Each iteration randomly samples 3--18 images per scene, with at most 18 images and four scenes in a batch. Training takes 35 hours on a single NVIDIA RTX PRO 6000 GPU.

\paragraph{Loss functions}

To improve training efficiency, we render the Gaussian scene at each of the $L$ levels and supervise it directly, without octree subdivision. Densification scores are supervised using gradients with respect to Gaussian positions.
The objective comprises ground-truth supervision, VFM alignment, and score prediction losses. Stochastic-solid volume rendering produces RGB images, depth maps, and normal maps for each level. Ground-truth supervision includes RGB L1 loss $\mathcal{L}_{\mathrm{RGB}}^{L1}$, RGB SSIM loss $\mathcal{L}_{\mathrm{RGB}}^{\mathrm{SSIM}}$, depth L1 loss $\mathcal{L}_{D}$, and normal cosine-consistency loss $\mathcal{L}_{N}$. Ground-truth normals are computed from depth-map gradients. VFM alignment comprises depth loss $\mathcal{L}_{D}^{\mathrm{VFM}}$ and normal loss $\mathcal{L}_{N}^{\mathrm{VFM}}$. These compare rendered depth and normals with the VFM's depth predictions and the derived normals.

The score loss follows the densification criterion of the original 3DGS\cite{kerbl3DGS}. During back-propagation, we accumulate the absolute gradients of projected Gaussian centers over multiple views at each level\cite{ye2024absgs}, as showin in \autoref{eq:accumulated_absolute_gradient}.
\begin{equation}
\label{eq:accumulated_absolute_gradient}
g_{v}^{(\ell)}
=\sum_{i=1}^{N}
\left(
\left|\frac{\partial\mathcal{L}_{i}^{(\ell)}}{\partial\mu_{i,v,x}^{\prime(\ell)}}\right|
+\left|\frac{\partial\mathcal{L}_{i}^{(\ell)}}{\partial\mu_{i,v,y}^{\prime(\ell)}}\right|
\right)
\end{equation}
Here, $v\in\mathcal{V}^{(\ell)}$ denotes a voxel at level $\ell$. The coordinates $\mu_{i,v,x}^{\prime(\ell)}$ and $\mu_{i,v,y}^{\prime(\ell)}$ denote its Gaussian's projected center in view $i$.
To stabilize optimization and reduce the influence of outliers, we transform the accumulated gradient using $\ln(1+\alpha_g g_{v}^{(\ell)})$, where $\alpha_g$ is a hyperparameter. We then apply the same sigmoid normalization as the score prediction head and compute an L1 score loss $\mathcal{L}_{S}$.
The complete training objective is shown in \autoref{eq:total_loss}.
\begin{equation}
\label{eq:total_loss}
\begin{aligned}
\mathcal{L}
={}(&\lambda_{\mathrm{RGB}}^{L1}\mathcal{L}_{\mathrm{RGB}}^{L1}
+\lambda_{\mathrm{RGB}}^{\mathrm{SSIM}}\mathcal{L}_{\mathrm{RGB}}^{\mathrm{SSIM}}
+\lambda_D\mathcal{L}_D
+\lambda_N\mathcal{L}_N)\\
&+(\lambda_D^{\mathrm{VFM}}\mathcal{L}_D^{\mathrm{VFM}}
+\lambda_N^{\mathrm{VFM}}\mathcal{L}_N^{\mathrm{VFM}})
+(\lambda_S\mathcal{L}_S)
\end{aligned}
\end{equation}
The default loss weights are $\lambda_{\mathrm{RGB}}^{L1}=7$, $\lambda_{\mathrm{RGB}}^{\mathrm{SSIM}}=3$, $\lambda_D=25$, $\lambda_N=0.2$, $\lambda_D^{\mathrm{VFM}}=25$, $\lambda_N^{\mathrm{VFM}}=0.2$, and $\lambda_S=1$.

\section{Experiments}
\subsection{RGB View Synthesis}
We evaluate NVS on Replica\cite{replica19arxiv}, TUM\cite{tum2012benchmark}, and Tanks and Temples\cite{TAT}. These datasets provide synthetic indoor, real-captured indoor, and real-captured outdoor scenes, respectively. We conducted qualitative and quantitative experiments between OctreeSplat and several SOTA baselines, which can be categorized into two groups: the first follows pixelSplat\cite{charatan2024pixelsplat} and requires camera parameters, including MVSplat\cite{chen2024mvsplat} and VolSplat\cite{wang2025volsplat}; the second uses VGGT-style\cite{wang2025vggt} reconstruction, including AnySplat\cite{jiang2025anysplat}, YoNoSplat\cite{ye2025yonosplatneedmodelfeedforward}, DA3\cite{depthanything3}, and QuerySplat\cite{li2026querysplat}. AnySplat and VolSplat use voxel alignment, whereas QuerySplat uses implicit query tokens. MVSplat, YoNoSplat, and DA3 use pixel alignment.
We measure RGB synthesis quality using PSNR, SSIM\cite{SSIM}, and LPIPS\cite{LPIPS}. These metrics characterize pixel-level fidelity, structural similarity, and perceptual similarity between synthesized and ground-truth views.

\begin{figure*}[!p]
    \centering
    \includegraphics[width=0.95\linewidth]{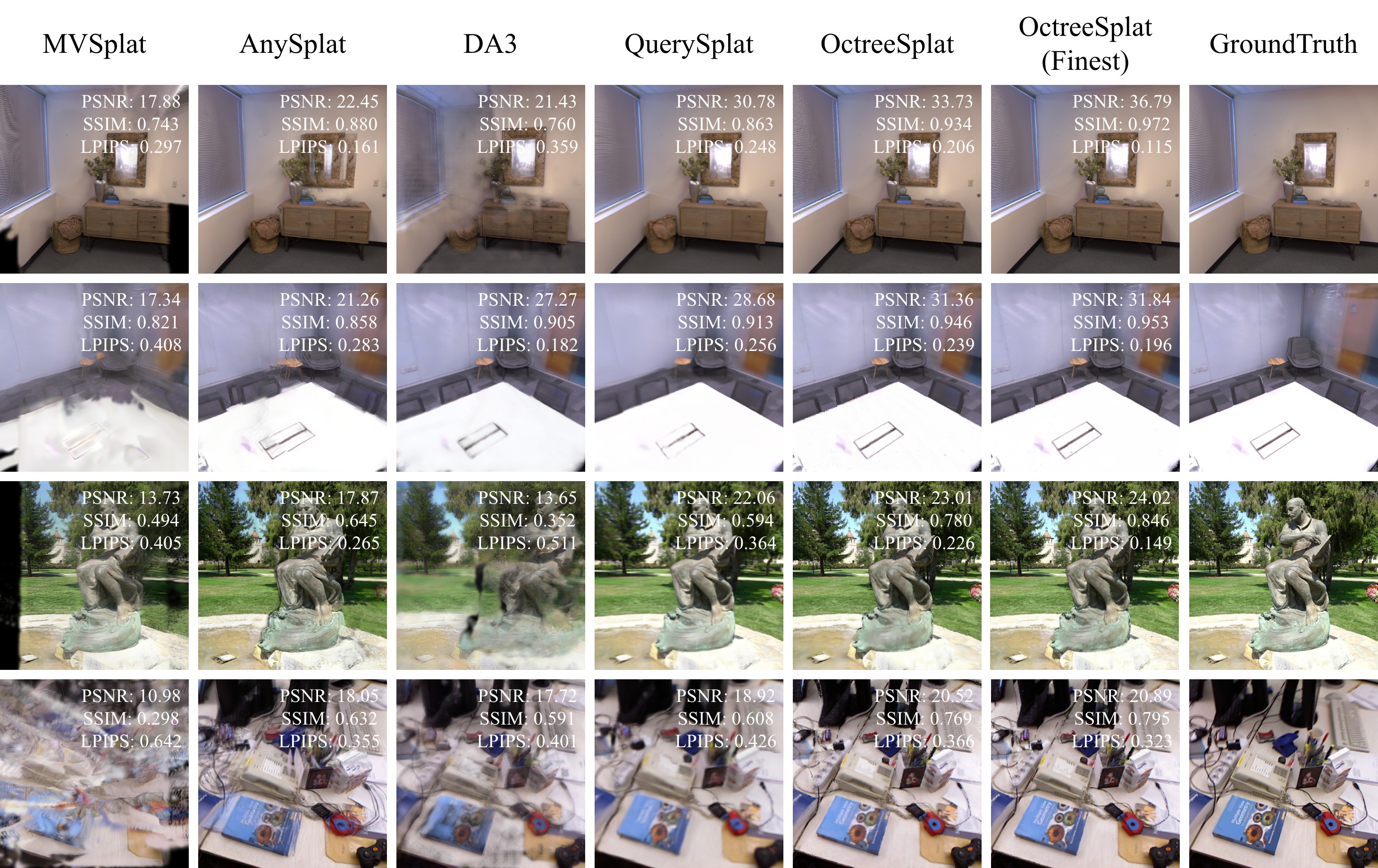}
    \caption{Qualitative comparison of RGB view synthesis.}
    \label{fig:nvs_expvis}
\par\vspace{1em}
    \begin{minipage}{\linewidth}
    \footnotesize
    \centering
    \captionof{table}{Quantitative RGB view synthesis results. Bold, underlined, and italic values indicate the best, second-best, and third-best results.}
    \label{tab:rgb_results}
    \setlength{\tabcolsep}{0pt}
    \renewcommand{\arraystretch}{1.0}
    \begin{tabularx}{\linewidth}{l*{12}{>{\centering\arraybackslash}X}}
        \toprule
        \multirow{2}{*}{Method} & \multicolumn{4}{c}{3 Views} & \multicolumn{4}{c}{10 Views} & \multicolumn{4}{c}{20 Views} \\
        \cmidrule(lr){2-5}\cmidrule(lr){6-9}\cmidrule(lr){10-13}
        & \shortstack{PSNR\\(dB) $\uparrow$} & \shortstack{SSIM\\$\uparrow$} & \shortstack{LPIPS\\$\downarrow$} & \shortstack{Count\\(100K)} & \shortstack{PSNR\\(dB) $\uparrow$} & \shortstack{SSIM\\$\uparrow$} & \shortstack{LPIPS\\$\downarrow$} & \shortstack{Count\\(100K)} & \shortstack{PSNR\\(dB) $\uparrow$} & \shortstack{SSIM\\$\uparrow$} & \shortstack{LPIPS\\$\downarrow$} & \shortstack{Count\\(100K)} \\
        \midrule
        \multicolumn{13}{l}{\textit{Replica}} \\
        \midrule
        MVSplat & 17.02 & 0.757 & 0.303 & 1.97 & 19.97 & 0.759 & 0.318 & 6.55 & 10.69 & 0.324 & 0.610 & 6.55 \\
        VolSplat & 19.83 & 0.754 & 0.450 & 1.97 & 16.15 & 0.667 & 0.471 & 6.54 & 14.77 & 0.615 & 0.567 & 6.53 \\
        AnySplat & 25.82 & 0.886 & \underline{0.159} & 6.74 & 23.25 & 0.825 & \underline{0.200} & 20.80 & 22.93 & 0.810 & 0.286 & 42.18 \\
        YoNoSplat & 25.68 & 0.833 & 0.279 & 1.51 & 24.52 & 0.809 & 0.262 & 5.02 & 23.65 & 0.806 & 0.331 & 10.04 \\
        DA3 & 21.42 & 0.782 & 0.335 & 8.04 & 23.64 & 0.837 & \textit{0.208} & 26.83 & 26.00 & \textit{0.860} & \underline{0.212} & 53.66 \\
        QuerySplat & \textit{33.14} & \textit{0.924} & 0.199 & 2.00 & \textit{30.03} & \textit{0.884} & 0.270 & 1.87 & \textit{28.32} & 0.859 & 0.316 & 1.75 \\
        OctreeSplat & \underline{34.27} & \underline{0.944} & \textit{0.198} & 2.81 & \underline{32.10} & \underline{0.923} & 0.228 & 8.32 & \underline{31.24} & \underline{0.910} & \textit{0.243} & 17.61 \\
        \begin{tabular}[c]{@{}c@{}}OctreeSplat(Finest)\end{tabular} & \textbf{36.22} & \textbf{0.966} & \textbf{0.128} & 7.32 & \textbf{33.47} & \textbf{0.947} & \textbf{0.158} & 22.83 & \textbf{32.40} & \textbf{0.934} & \textbf{0.185} & 46.08 \\
        \midrule
        \multicolumn{13}{l}{\textit{Tanks and Temples}} \\
        \midrule
        MVSplat & 14.18 & 0.537 & 0.401 & 1.97 & 15.65 & 0.541 & 0.420 & 6.55 & 13.76 & 0.431 & 0.544 & 6.55 \\
        VolSplat & 15.28 & 0.520 & 0.506 & 1.96 & 14.98 & 0.544 & 0.439 & 6.55 & 12.87 & 0.356 & 0.595 & 6.53 \\
        AnySplat & 19.24 & 0.715 & \textit{0.263} & 6.14 & 16.89 & \textit{0.615} & \textit{0.318} & 19.71 & 16.67 & 0.536 & \textit{0.394} & 29.35 \\
        YoNoSplat & 12.30 & 0.372 & 0.588 & 1.51 & 13.17 & 0.420 & 0.533 & 5.02 & 13.62 & 0.374 & 0.587 & 10.04 \\
        DA3 & 17.63 & 0.551 & 0.398 & 8.04 & 18.26 & 0.567 & 0.373 & 26.83 & 18.48 & 0.533 & 0.428 & 53.66 \\
        QuerySplat & \underline{25.43} & \textit{0.745} & 0.286 & 2.34 & \underline{22.31} & 0.612 & 0.397 & 2.00 & \textbf{22.17} & \textit{0.647} & 0.397 & 2.01 \\
        OctreeSplat & \textit{25.10} & \underline{0.826} & \underline{0.253} & 4.73 & \textit{22.28} & \underline{0.746} & \underline{0.306} & 13.97 & \textit{20.52} & \underline{0.662} & \underline{0.385} & 20.58 \\
        \begin{tabular}[c]{@{}c@{}}OctreeSplat(Finest)\end{tabular} & \textbf{26.51} & \textbf{0.876} & \textbf{0.170} & 7.22 & \textbf{22.77} & \textbf{0.784} & \textbf{0.253} & 22.79 & \underline{21.32} & \textbf{0.697} & \textbf{0.336} & 39.68 \\
        \midrule
        \multicolumn{13}{l}{\textit{TUM}} \\
        \midrule
        MVSplat & 11.03 & 0.412 & 0.551 & 1.97 & 12.01 & 0.407 & 0.585 & 6.55 & 11.70 & 0.386 & 0.617 & 6.55 \\
        VolSplat & 14.61 & 0.466 & 0.572 & 1.97 & 13.61 & 0.447 & 0.582 & 6.55 & 13.16 & 0.423 & 0.610 & 6.54 \\
        AnySplat & 24.33 & \textit{0.825} & \underline{0.203} & 5.38 & 19.20 & \textit{0.707} & \underline{0.315} & 14.00 & 18.00 & \textit{0.655} & \underline{0.365} & 28.33 \\
        YoNoSplat & 16.55 & 0.552 & 0.471 & 1.51 & 16.36 & 0.534 & 0.474 & 5.02 & 16.01 & 0.530 & 0.504 & 10.04 \\
        DA3 & 19.62 & 0.692 & 0.363 & 8.04 & 16.51 & 0.626 & 0.439 & 26.83 & 17.63 & 0.639 & 0.423 & 53.66 \\
        QuerySplat & \textit{24.61} & 0.791 & 0.299 & 1.66 & \textit{20.53} & 0.682 & 0.414 & 1.30 & \textit{19.71} & 0.650 & 0.459 & 1.13 \\
        OctreeSplat & \underline{27.86} & \underline{0.858} & \textit{0.282} & 3.48 & \textbf{21.86} & \underline{0.767} & \textit{0.364} & 10.40 & \underline{21.00} & \underline{0.723} & \textit{0.394} & 17.46 \\
        \begin{tabular}[c]{@{}c@{}}OctreeSplat(Finest)\end{tabular} & \textbf{29.40} & \textbf{0.893} & \textbf{0.202} & 7.33 & \underline{21.83} & \textbf{0.775} & \textbf{0.307} & 23.98 & \textbf{21.38} & \textbf{0.733} & \textbf{0.354} & 43.49 \\
        \bottomrule
    \end{tabularx}
    \end{minipage}
\end{figure*}

As shown in \autoref{tab:rgb_results} and \autoref{fig:nvs_expvis}, OctreeSplat produces clear RGB renderings and suppresses overlap artifacts. OctreeSplat (Finest) directly uses Gaussian primitives predicted from the finest voxel level.
OctreeSplat efficiently controls the primitive count, which reduces the count by approximately half or more in most settings, with a quality--compactness trade-off relative to the finest-level variant.

\subsection{Depth Map Synthesis}
We evaluate depth synthesis using the same datasets and baselines as for RGB synthesis. Metrics include root mean square error (RMSE), absolute relative error (AbsRel), and threshold accuracy ($\delta_m$). RMSE measures the root mean squared difference between predicted depth $\hat d$ and ground-truth depth $d$. For $N_{\mathrm{valid}}$ valid pixels, AbsRel is $\mathrm{AbsRel}=\frac{1}{N_{\mathrm{valid}}}\sum_{i=1}^{N_{\mathrm{valid}}}\frac{|\hat d_i-d_i|}{d_i}$. Threshold accuracy is the fraction of pixels satisfying $\max(\hat d_i/d_i,d_i/\hat d_i)<m$, with $m=1.05$ and $m=1.10$.

\begin{figure*}[!p]
    \centering
    \includegraphics[width=0.95\linewidth]{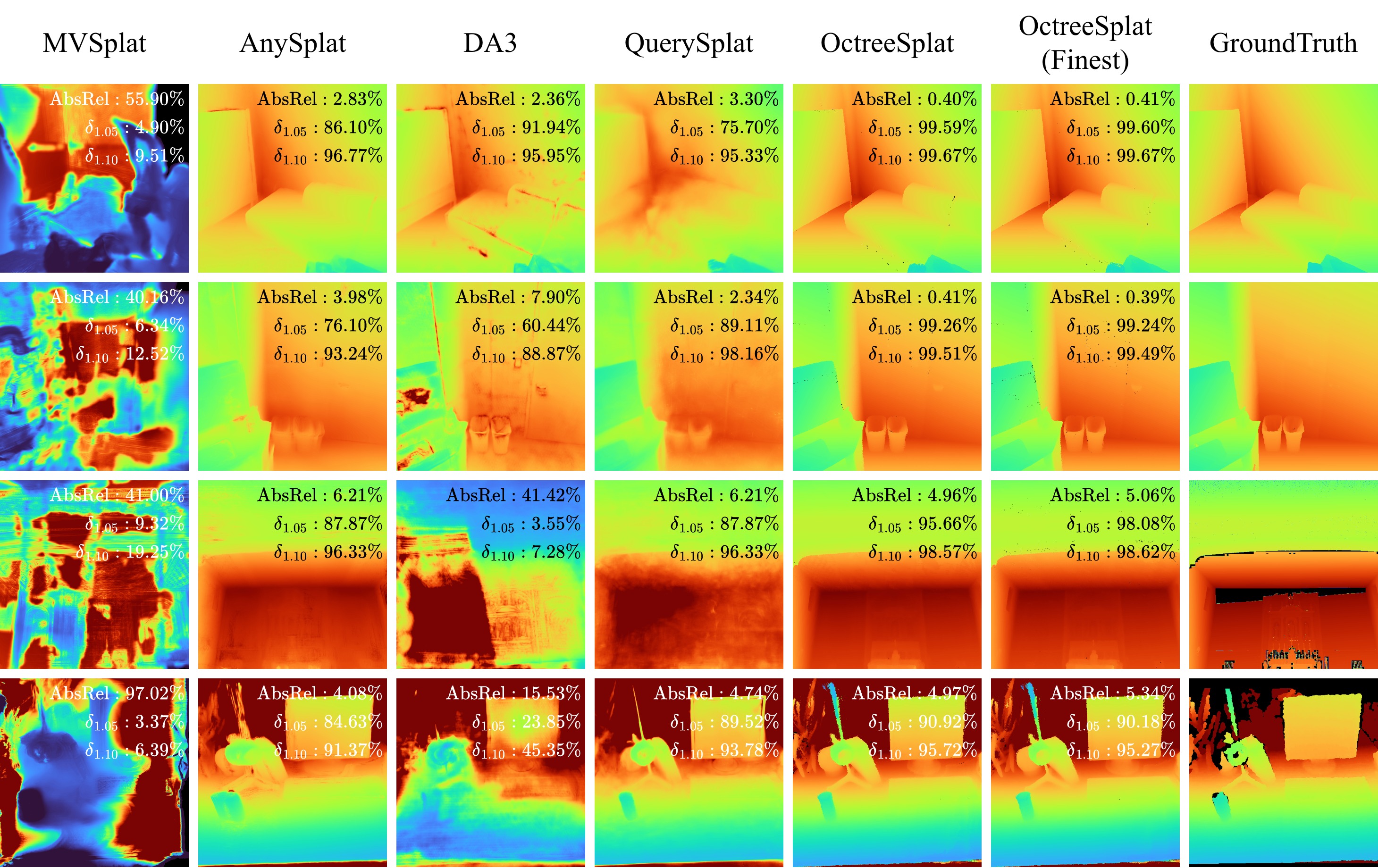}
    \caption{Qualitative comparison of depth map synthesis.}
    \label{fig:depth_expvis}
\par\vspace{1em}
\begin{minipage}{\linewidth}
    \footnotesize
    \centering
    \captionof{table}{Quantitative depth synthesis results.}
    \label{tab:geometry_results}
    \setlength{\tabcolsep}{0pt}
    \renewcommand{\arraystretch}{1.0}
    \footnotesize
    \begin{tabularx}{\linewidth}{l*{12}{>{\centering\arraybackslash}X}}
        \toprule
        \multirow{2}{*}{Method} & \multicolumn{4}{c}{3 Views} & \multicolumn{4}{c}{10 Views} & \multicolumn{4}{c}{20 Views} \\
        \cmidrule(lr){2-5}\cmidrule(lr){6-9}\cmidrule(lr){10-13}
         & \shortstack{RMSE\\(cm) $\downarrow$} & \shortstack{AbsRel\\(\%) $\downarrow$} & \shortstack{$\delta_{1.05}$\\$\uparrow$} & \shortstack{$\delta_{1.10}$\\$\uparrow$} & \shortstack{RMSE\\(cm) $\downarrow$} & \shortstack{AbsRel\\(\%) $\downarrow$} & \shortstack{$\delta_{1.05}$\\$\uparrow$} & \shortstack{$\delta_{1.10}$\\$\uparrow$} & \shortstack{RMSE\\(cm) $\downarrow$} & \shortstack{AbsRel\\(\%) $\downarrow$} & \shortstack{$\delta_{1.05}$\\$\uparrow$} & \shortstack{$\delta_{1.10}$\\$\uparrow$} \\
        \midrule
        \multicolumn{13}{l}{\textit{Replica}} \\
        \midrule
        MVSplat & 114.92 & 30.90 & 12.78 & 24.15 & 216.08 & 69.27 & 1.98 & 3.89 & 202.49 & 66.09 & 2.26 & 4.37 \\
        VolSplat & 72.77 & 20.89 & 16.19 & 30.36 & 102.87 & 30.43 & 11.88 & 22.56 & 100.54 & 31.53 & 12.40 & 23.16 \\
        AnySplat & 15.39 & 3.51 & \textit{76.32} & \textit{95.70} & 33.73 & 9.58 & 59.05 & 76.88 & 21.63 & 5.52 & 65.52 & 86.30 \\
        YoNoSplat & 27.04 & 7.59 & 44.76 & 73.48 & 23.47 & 7.28 & 51.08 & 75.58 & 22.64 & 7.34 & 51.25 & 76.38 \\
        DA3 & 105.30 & 32.62 & 13.49 & 25.20 & 125.53 & 43.28 & 28.31 & 46.40 & 21.39 & 5.65 & 73.96 & 88.62 \\
        QuerySplat & \textit{7.25} & \textit{1.66} & \textit{95.30} & \textit{98.98} & \textit{8.88} & \textit{2.36} & \textit{89.25} & \textit{96.18} & \textit{10.93} & \textit{2.94} & \textit{84.26} & \textit{95.20} \\
        OctreeSplat & \underline{4.50} & \underline{0.38} & \textbf{99.50} & \textbf{99.71} & \underline{5.35} & \underline{0.48} & \underline{99.31} & \underline{99.58} & \underline{6.25} & \underline{0.59} & \underline{99.04} & \underline{99.40} \\
        \begin{tabular}[c]{@{}c@{}}OctreeSplat(Finest)\end{tabular} & \textbf{4.37} & \textbf{0.36} & \underline{99.50} & \underline{99.71} & \textbf{5.19} & \textbf{0.45} & \textbf{99.34} & \textbf{99.60} & \textbf{6.00} & \textbf{0.54} & \textbf{99.17} & \textbf{99.45} \\
        \midrule
        
\multicolumn{13}{l}{\textit{Tanks and Temples}} \\
        \midrule
        MVSplat & 486.96 & 35.15 & 12.96 & 24.89 & 708.32 & 64.97 & 0.97 & 1.95 & 880.57 & 63.64 & 10.82 & 20.12 \\
        VolSplat & 402.38 & 35.34 & 8.76 & 16.85 & 404.29 & 37.41 & 8.04 & 15.44 & 409.49 & 40.08 & 10.61 & 21.12 \\
        AnySplat & \textit{111.23} & 5.13 & 67.28 & \underline{89.96} & \textit{91.41} & \textit{4.19} & 74.78 & \textit{92.55} & \textbf{83.96} & \textit{8.76} & \textit{81.63} & \textit{91.20} \\
        YoNoSplat & 416.84 & 27.57 & 8.02 & 17.22 & 423.07 & 40.87 & 6.45 & 12.69 & 197.36 & 27.17 & 28.12 & 41.91 \\
        DA3 & 416.05 & 31.97 & 15.11 & 27.21 & 368.92 & 38.28 & 15.65 & 30.66 & 255.79 & 26.38 & 30.98 & 42.56 \\
        QuerySplat & 135.41 & \underline{4.53} & \textbf{78.10} & \textbf{90.39} & 190.17 & 4.99 & \textit{82.93} & 92.11 & 325.92 & 24.42 & 59.00 & 75.41 \\
        OctreeSplat & \underline{91.48} & \textit{4.81} & \textit{74.09} & 85.56 & \underline{65.89} & \underline{2.22} & \underline{93.01} & \underline{97.38} & \textit{114.33} & \underline{8.74} & \underline{92.14} & \underline{94.61} \\
        \begin{tabular}[c]{@{}c@{}}OctreeSplat(Finest)\end{tabular} & \textbf{87.77} & \textbf{3.56} & \underline{76.88} & \textit{89.34} & \textbf{54.22} & \textbf{1.78} & \textbf{95.36} & \textbf{98.45} & \underline{104.48} & \textbf{8.33} & \textbf{93.37} & \textbf{94.73} \\
        \midrule
        \multicolumn{13}{l}{\textit{TUM}} \\
        \midrule
        MVSplat & 182.10 & 16.35 & 24.51 & 45.02 & 1388.26 & 105.96 & 6.51 & 12.66 & 1292.47 & 95.65 & 4.63 & 9.03 \\
        VolSplat & 162.56 & 13.74 & 25.96 & 48.93 & 146.04 & 13.05 & 30.59 & 50.66 & 187.98 & 13.59 & 34.41 & 55.10 \\
        AnySplat & \textit{70.27} & \textit{5.00} & \textit{65.57} & \textit{83.67} & \textit{77.69} & \textit{5.44} & 66.31 & \textit{85.03} & \textit{113.17} & \textit{6.32} & \textit{69.34} & \textit{86.30} \\
        YoNoSplat & 117.50 & 8.71 & 42.38 & 66.10 & 104.08 & 7.91 & 57.47 & 76.95 & 120.41 & 8.65 & 54.82 & 74.42 \\
        DA3 & 150.90 & 16.08 & 19.67 & 39.67 & 166.85 & 15.20 & 30.89 & 55.73 & 173.82 & 11.60 & 47.10 & 68.50 \\
        QuerySplat & 91.88 & 7.18 & 62.08 & 81.05 & 97.82 & 8.49 & \textit{68.53} & 81.99 & 114.56 & 9.23 & 60.39 & 74.53 \\
        OctreeSplat & \textbf{51.35} & \textbf{2.28} & \textbf{92.99} & \textbf{96.62} & \textbf{61.73} & \textbf{2.77} & \underline{91.95} & \textbf{96.48} & \textbf{76.98} & \textbf{3.11} & \textbf{90.41} & \textbf{95.67} \\
        \begin{tabular}[c]{@{}c@{}}OctreeSplat(Finest)\end{tabular} & \underline{51.67} & \underline{2.29} & \underline{92.95} & \underline{96.60} & \underline{64.89} & \underline{2.87} & \textbf{91.96} & \underline{96.35} & \underline{79.54} & \underline{3.20} & \underline{90.31} & \underline{95.62} \\
        \bottomrule
    \end{tabularx}
\end{minipage}
\end{figure*}

As shown in \autoref{tab:geometry_results} and \autoref{fig:depth_expvis}, OctreeSplat shows strong depth synthesis performance across the evaluated datasets.

\subsection{Pose Estimation}
OctreeSplat inherits the pretrained DA3 backbone and therefore produces the same pose predictions before TTO. When ground-truth camera parameters are available, the additional TTO module improves alignment between predictions and the ground-truth parameters.
TTO uses 20 iterations with a lr of $5\times10^{-4}$.
As shown in \autoref{tab:pose_results}, OctreeSplat and DA3 have identical pose accuracy before adaptation. Conditioning DA3 on camera parameters, denoted DA3 (Pose), improves poses accuracy. OctreeSplat does not encode camera parameters into the backbone. Instead, TTO uses them as supervision and achieves lower translation and rotation errors, with particularly clear gains in rotation accuracy.

\begin{table}[!htbp]
    \centering
    \caption{Quantitative pose estimation results with 20 input views.}
    \label{tab:pose_results}
    \setlength{\tabcolsep}{2pt}
    \renewcommand{\arraystretch}{0.9}
    \footnotesize
    \begin{tabularx}{\linewidth}{l*{4}{>{\centering\arraybackslash}X}}
        \toprule
        \multirow{2}{*}{Method} & \multicolumn{2}{c}{Trans. (mm) $\downarrow$} & \multicolumn{2}{c}{Rot. (${}^\circ$) $\downarrow$} \\
        \cmidrule(lr){2-3}\cmidrule(lr){4-5}
        & Mean & Median & Mean & Median \\
        \midrule
        \multicolumn{5}{l}{\textit{DTU}} \\
        \midrule
        YoNoSplat & 16.383 & 11.687 & 5.562 & 5.210 \\
        AnySplat & 5.846 & 4.581 & 2.105 & 2.036 \\
        QuerySplat & 3.043 & 2.773 & 0.971 & 0.969 \\
        OctreeSplat \& DA3 & 2.007 & 1.535 & 0.530 & 0.511 \\
        DA3 (Pose) & 1.250 & 1.049 & 0.503 & 0.477 \\
        \shortstack{OctreeSplat (TTO)} & 1.134 & 0.903 & 0.254 & 0.238 \\
        \midrule
        \multicolumn{5}{l}{\textit{Replica}} \\
        \midrule
        YoNoSplat & 118.385 & 70.258 & 5.195 & 4.415 \\
        AnySplat & 109.801 & 95.020 & 13.474 & 13.754 \\
        QuerySplat & 15.357 & 12.439 & 0.825 & 0.786 \\
        OctreeSplat \& DA3 & 8.434 & 5.981 & 0.444 & 0.400 \\
        DA3 (Pose) & 6.008 & 4.019 & 0.499 & 0.450 \\
        \shortstack{OctreeSplat (TTO)} & 4.045 & 3.172 & 0.205 & 0.192 \\
        \bottomrule
    \end{tabularx} 
\end{table}

\subsection{Surface Reconstruction}
Following the DTU evaluation protocol\cite{dtu}, we measure surface reconstruction quality using the symmetric Chamfer distance (CD). Let $P$ and $Q$ denote points sampled from the reconstructed and ground-truth surfaces. Accuracy (Acc.) and completeness (Com.) are the symmetric mean nearest-neighbor distances, and CD is their arithmetic mean, as shown in \autoref{eq:chamfer_distance}.
\begin{equation}
\label{eq:chamfer_distance}
\begin{aligned}
\mathrm{CD}(P,Q)&=\frac{\mathrm{Acc.}(P,Q)+\mathrm{Com.}(P,Q)}{2}\\
&=\frac{1}{2}\Biggl(
\frac{1}{|P|}\sum_{\mathbf{p}\in P}\min_{\mathbf{q}\in Q}\|\mathbf{p}-\mathbf{q}\|_2\\
&\qquad+\frac{1}{|Q|}\sum_{\mathbf{q}\in Q}\min_{\mathbf{p}\in P}\|\mathbf{q}-\mathbf{p}\|_2\Biggr).
\end{aligned}
\end{equation}

Because OctreeSplat and the evaluated baselines reconstruct scenes without input poses, we first align predicted and ground-truth camera poses using the Umeyama algorithm\cite{umeyama1991least}. Joint errors in poses, intrinsics, and depth can leave residual surface misalignment after camera alignment. We therefore also apply iterative closest point (ICP) alignment between reconstructed and ground-truth surfaces before computing CD. Results before and after ICP are reported separately to distinguish alignment errors from residual surface discrepancies.

\begin{figure}[!htbp]
    \centering
    \includegraphics[width=\linewidth]{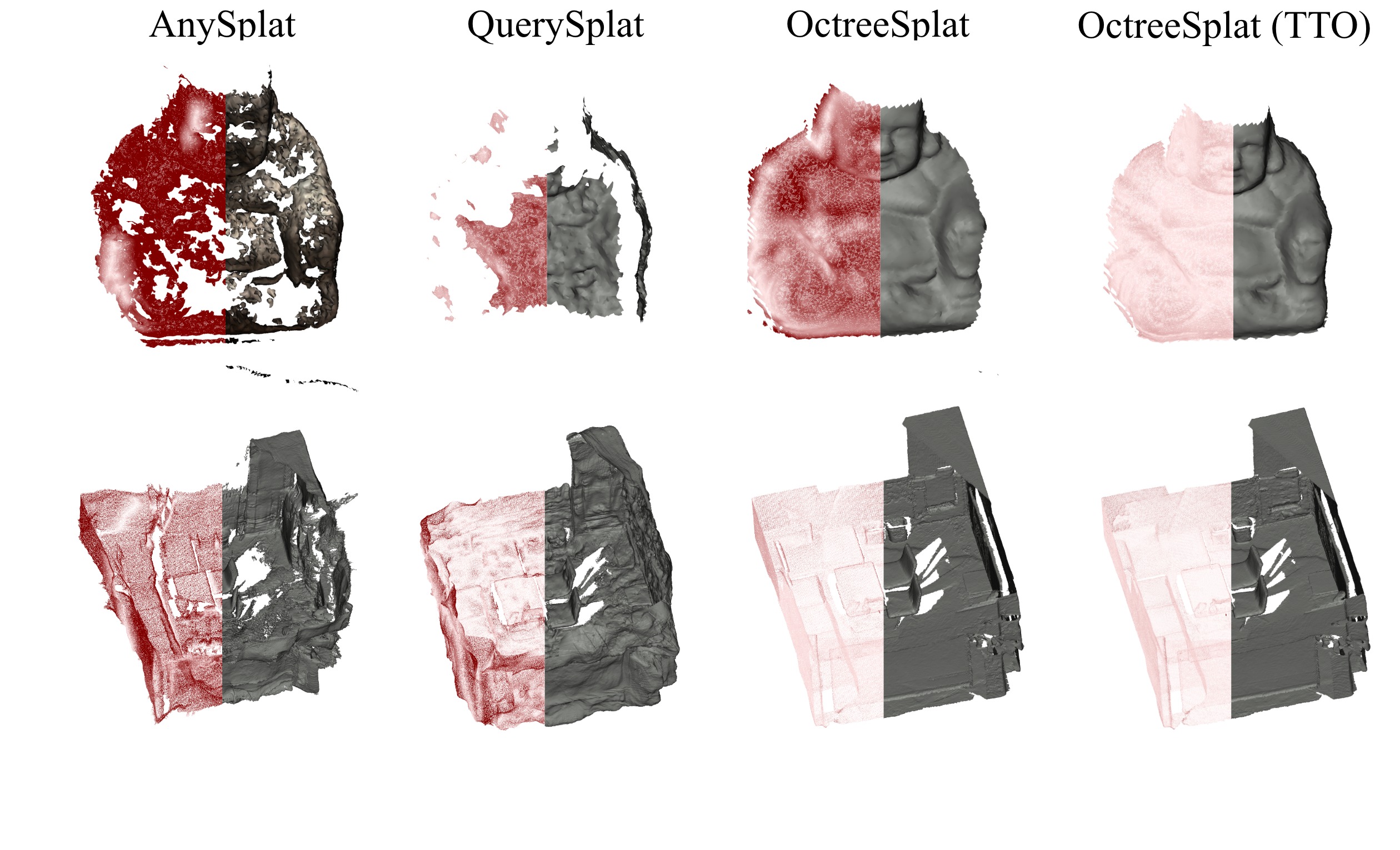}
    \caption{Qualitative comparison of reconstructed surfaces.}
    \label{fig:mesh_vis}
\end{figure}

As shown in \autoref{tab:icp_geometry_results} and \autoref{fig:mesh_vis}, OctreeSplat improves surface reconstruction. Adding TTO reduces surface misalignment caused by pose errors and further improves the reconstruction accuracy after ICP.

\begin{table}[htbp]
    \centering
    \caption{Surface reconstruction results before and after ICP alignment. All distances are measured in mm.}
    \label{tab:icp_geometry_results}
    \setlength{\tabcolsep}{2pt}
    \renewcommand{\arraystretch}{0.9}
    \begin{tabularx}{\linewidth}{l*{6}{>{\centering\arraybackslash}X}}
        \toprule
        \multirow{2}{*}{Method} & \multicolumn{3}{c}{Before ICP} & \multicolumn{3}{c}{After ICP} \\
        \cmidrule(lr){2-4}\cmidrule(lr){5-7}
        & Acc. & Com. & CD & Acc. & Com. & CD \\
        \midrule
        \multicolumn{7}{l}{\textit{Replica}} \\
        \midrule
        AnySplat & 51.905 & 49.643 & 50.774 & 41.793 & 38.600 & 40.196 \\
        DA3 & 65.833 & 64.426 & 65.129 & 46.325 & 43.459 & 44.892 \\
        YoNoSplat & 48.781 & 46.121 & 47.451 & 43.176 & 40.751 & 41.963 \\
        QuerySplat & 36.569 & 34.523 & 35.546 & 33.769 & 31.730 & 32.750 \\
        OctreeSplat & 15.108 & 14.504 & 14.806 & 12.105 & 11.686 & 11.895 \\
        \shortstack{OctreeSplat (TTO)} & 9.221 & 8.831 & 9.026 & 8.495 & 8.172 & 8.333 \\
        \midrule
        \multicolumn{7}{l}{\textit{DTU}} \\
        \midrule
        AnySplat & 4.408 & 11.186 & 7.797 & 3.544 & 9.976 & 6.760 \\
        DA3 &  & Fail& & & Fail& \\
        YoNoSplat & 7.740 & 12.830 & 10.285 & 7.287 & 12.020 & 9.654 \\
        QuerySplat & 5.194 & 9.382 & 7.288 & 2.740 & 5.757 & 4.249 \\
        OctreeSplat & 7.703 & 9.337 & 8.520 & 3.035 & 3.344 & 3.190 \\
        \shortstack{OctreeSplat (TTO)} & 3.187 & 5.700 & 4.443 & 1.105 & 1.914 & 1.510 \\
        \bottomrule
    \end{tabularx}
\end{table}

\subsection{Efficiency}
\begin{figure}[!ht]
    \centering
    \includegraphics[width=0.95\linewidth]{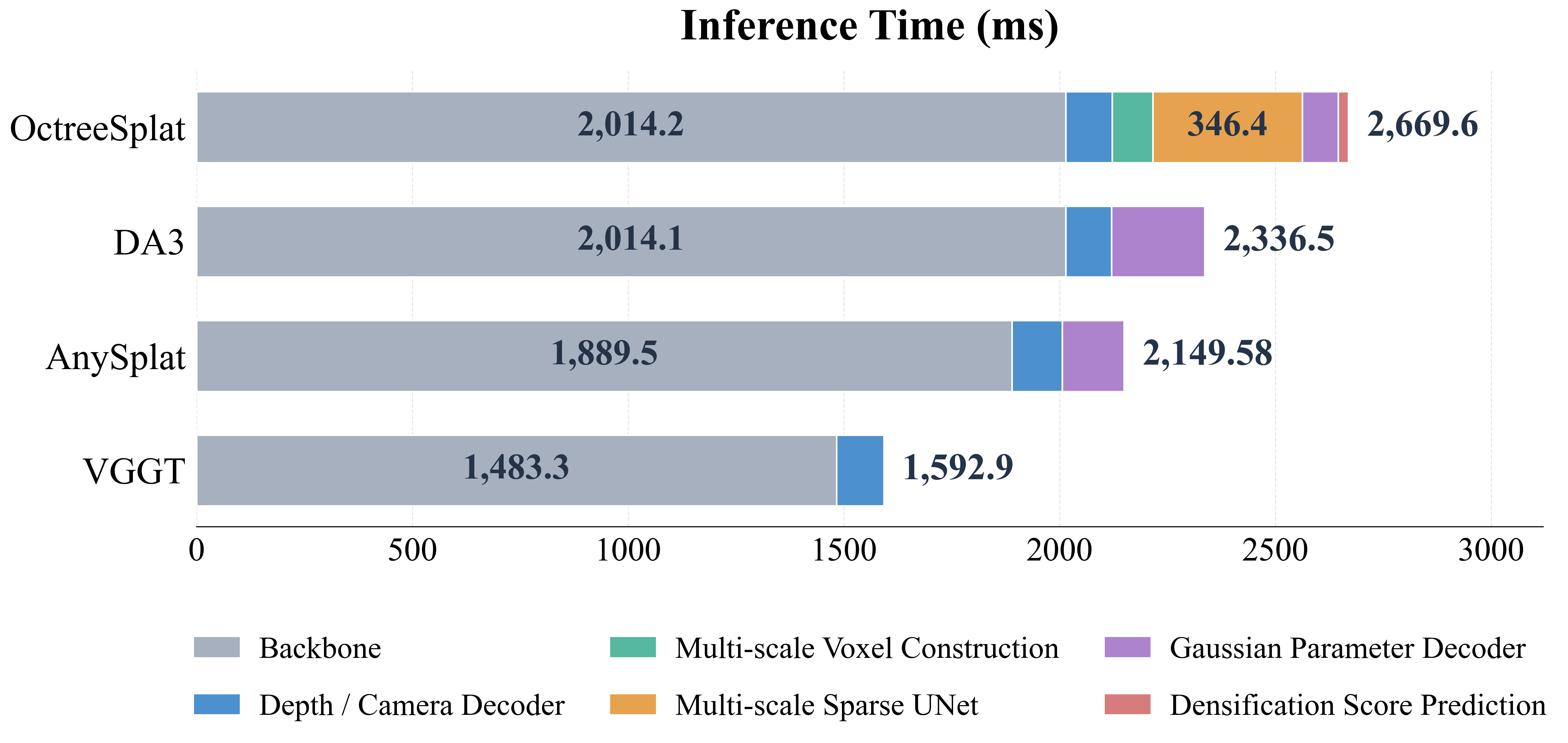}
    \caption{Model inference time.}
    \label{fig:inference_time}
\end{figure}

The inference time of OctreeSplat is shown in \autoref{fig:inference_time}. The total inference time is 2.66 s, compared with 2.33 s for DA3 using the same backbone, an increase of 14.2\%. Most additional computation comes from the multi-scale SPUNet.

When TTO is enabled, OctreeSplat runs the VFM backbone in bf16 to improve efficiency, with little effect on accuracy. A backbone forward pass takes 287.9 ms, and back-propagation takes 426.0 ms. The total runtime with TTO remains below 15 s.

\subsection{Ablation Studies}
\subsubsection{Stochastic-Solid Volume Rendering}

We first replace stochastic-solid volume rendering with the vanilla 3DGS renderer\cite{kerbl3DGS}. As shown in \autoref{tab:ablation_studies}(a), after replacement, RGB synthesis accuracy shows a slight decline, while the depth synthesis accuracy exhibits a significant degradation. This confirms the effectiveness of the stochastic solid-volume rendering module.

\begin{table*}[!h]
    \centering
    \captionsetup{position=top}
    \captionsetup[subtable]{position=top}
    \caption{Ablation studies. (a) Stochastic-solid volume rendering. (b) Depth supervision. (c) Voxel size. (d) Octree subdivision.}
    \label{tab:ablation_studies}
    \setlength{\tabcolsep}{1pt}
    \renewcommand{\arraystretch}{1.2}
    \footnotesize
    
    \settowidth{\ablationdatasetwidth}{Dataset}
    \addtolength{\ablationdatasetwidth}{18pt}
    \settowidth{\ablationmethodwidth}{OctreeSplat (w/o depth losses)}
    \addtolength{\ablationmethodwidth}{0pt}
    \subfloat[Stochastic-solid volume rendering\label{tab:volrender_ablation}]{%
        \begin{minipage}{\linewidth}
        \centering\footnotesize
        \begin{tabularx}{\linewidth}{>{\raggedright\arraybackslash}p{\ablationdatasetwidth}>{\raggedright\arraybackslash}p{\ablationmethodwidth}*{7}{>{\centering\arraybackslash}X}}
        \toprule
        Dataset & Method & PSNR $\uparrow$ & SSIM $\uparrow$ & LPIPS $\downarrow$ & RMSE $\downarrow$ & AbsRel $\downarrow$ & $\delta_{1.05}$ $\uparrow$ & $\delta_{1.10}$ $\uparrow$ \\
        \midrule
        \multirow{2}{*}{Replica} & OctreeSplat & 32.10 & 0.923 & 0.228 & 5.35 & 0.48 & 99.31 & 99.58 \\
        & \mbox{OctreeSplat(3DGS)} & 29.11 & 0.889 & 0.213 & 9.58 & 1.01 & 96.09 & 97.92 \\
        \midrule
        \multirow{2}{*}{TUM} & OctreeSplat & 21.86 & 0.767 & 0.364 & 61.73 & 2.77 & 91.95 & 96.48 \\
        & \mbox{OctreeSplat(3DGS)} & 20.68 & 0.727 & 0.357 & 286.14 & 3.59 & 91.16 & 96.16 \\
        \bottomrule
    \end{tabularx}
        \end{minipage}%
    }
    \par\vspace{4pt}
    \subfloat[Depth supervision\label{tab:depth_ablation}]{%
        \begin{minipage}{\linewidth}
        \centering\footnotesize
        \begin{tabularx}{\linewidth}{>{\raggedright\arraybackslash}p{\ablationdatasetwidth}>{\raggedright\arraybackslash}p{\ablationmethodwidth}*{7}{>{\centering\arraybackslash}X}}
        \toprule
        Dataset & Method & PSNR $\uparrow$ & SSIM $\uparrow$ & LPIPS $\downarrow$ & RMSE $\downarrow$ & AbsRel $\downarrow$ & $\delta_{1.05}$ $\uparrow$ & $\delta_{1.10}$ $\uparrow$ \\
        \midrule
        \multirow{3}{*}{Replica} & OctreeSplat & 32.10 & 0.923 & 0.228 & 5.35 & 0.48 & 99.31 & 99.58 \\
         & \mbox{OctreeSplat (w/o data depth)} & 32.56 & 0.928 & 0.219 & 5.39 & 0.48 & 99.29 & 99.58 \\
         & \mbox{OctreeSplat (w/o depth losses)} & 33.42 & 0.935 & 0.216 & 5.26 & 0.57 & 99.18 & 99.60 \\
        \midrule
        \multirow{3}{*}{TUM} & OctreeSplat & 21.86 & 0.767 & 0.364 & 61.73 & 2.77 & 91.95 & 96.48 \\
         & \mbox{OctreeSplat (w/o data depth)} & 21.80 & 0.772 & 0.353 & 62.70 & 2.83 & 91.94 & 96.38 \\
         & \mbox{OctreeSplat (w/o depth losses)} & 23.19 & 0.807 & 0.344 & 64.82 & 2.86 & 92.17 & 96.65 \\
        \bottomrule
    \end{tabularx}
        \end{minipage}%
    }
    \par\vspace{4pt}
    \subfloat[Voxel size\label{tab:voxel_size_ablation}]{%
        \begin{minipage}{\linewidth}
        \centering\footnotesize
        \begin{tabularx}{\linewidth}{>{\raggedright\arraybackslash}p{\ablationdatasetwidth}>{\raggedright\arraybackslash}p{\ablationmethodwidth}*{7}{>{\centering\arraybackslash}X}c}
        \toprule
        Dataset & Method & PSNR $\uparrow$ & SSIM $\uparrow$ & LPIPS $\downarrow$ & RMSE $\downarrow$ & AbsRel $\downarrow$ & $\delta_{1.05}$ $\uparrow$ & $\delta_{1.10}$ $\uparrow$ & \mbox{Count(100K) $\downarrow$} \\
        \midrule
        \multirow{6}{*}{Replica} & \mbox{OctreeSplat (6mm)} & 32.73 & 0.931 & 0.206 & 5.21 & 0.48 & 99.33 & 99.60 & 11.85 \\
         & \mbox{OctreeSplat (8mm)} & 32.10 & 0.923 & 0.228 & 5.35 & 0.48 & 99.31 & 99.58 & 8.32 \\
         & \mbox{OctreeSplat (12mm)} & 30.57 & 0.904 & 0.275 & 5.74 & 0.49 & 99.22 & 99.52 & 4.64 \\
         & \mbox{OctreeSplat (16mm)} & 28.98 & 0.883 & 0.319 & 6.11 & 0.52 & 99.12 & 99.44 & 2.86 \\
         & \mbox{OctreeSplat (32mm)} & 21.16 & 0.784 & 0.465 & 9.31 & 0.98 & 98.26 & 98.81 & 0.74 \\
         & \mbox{OctreeSplat (64mm)} & 15.70 & 0.701 & 0.582 & 15.79 & 2.74 & 94.88 & 96.30 & 0.17 \\
        \midrule
        \multirow{6}{*}{TUM} & \mbox{OctreeSplat (6mm)} & 22.12 & 0.778 & 0.349 & 63.42 & 2.84 & 91.94 & 96.45 & 13.69 \\
         & \mbox{OctreeSplat (8mm)} & 21.86 & 0.767 & 0.364 & 61.73 & 2.77 & 91.95 & 96.48 & 10.40 \\
         & \mbox{OctreeSplat (12mm)} & 21.30 & 0.746 & 0.388 & 58.23 & 2.67 & 91.94 & 96.48 & 6.43 \\
         & \mbox{OctreeSplat (16mm)} & 20.62 & 0.724 & 0.417 & 56.32 & 2.63 & 91.83 & 96.44 & 4.25 \\
         & \mbox{OctreeSplat (32mm)} & 17.51 & 0.645 & 0.508 & 55.17 & 2.67 & 90.82 & 95.88 & 1.29 \\
         & \mbox{OctreeSplat (64mm)} & 13.27 & 0.560 & 0.608 & 64.50 & 3.18 & 87.52 & 93.85 & 0.35 \\
        \bottomrule
    \end{tabularx}
        \end{minipage}%
    }
    \par\vspace{4pt}
    \subfloat[Octree subdivision\label{tab:octree_split_ablation}]{%
        \begin{minipage}{\linewidth}
        \centering\footnotesize
        \begin{tabularx}{\linewidth}{>{\raggedright\arraybackslash}p{\ablationdatasetwidth}>{\raggedright\arraybackslash}p{\ablationmethodwidth}*{7}{>{\centering\arraybackslash}X}c}
        \toprule
        Dataset & Method & PSNR $\uparrow$ & SSIM $\uparrow$ & LPIPS $\downarrow$ & RMSE $\downarrow$ & AbsRel $\downarrow$ & $\delta_{1.05}$ $\uparrow$ & $\delta_{1.10}$ $\uparrow$ & \mbox{Count(100K) $\downarrow$} \\
        \midrule
        \multirow{4}{*}{Replica} & OctreeSplat & 32.10 & 0.923 & 0.228 & 5.35 & 0.48 & 99.31 & 99.58 & 8.32 \\
        & \mbox{OctreeSplat (Grad)} & 32.40 & 0.926 & 0.227 & 5.29 & 0.48 & 99.32 & 99.59 & 8.39 \\
        & \mbox{OctreeSplat (Point)} & 30.89 & 0.913 & 0.214 & 5.57 & 0.48 & 99.25 & 99.54 & 12.22 \\
        & \mbox{OctreeSplat (Finest)} & 33.47 & 0.947 & 0.158 & 5.19 & 0.45 & 99.34 & 99.60 & 22.83 \\
        \midrule
        \multirow{4}{*}{TUM} & OctreeSplat & 21.86 & 0.767 & 0.364 & 61.73 & 2.77 & 91.95 & 96.48 & 10.40 \\
        & \mbox{OctreeSplat (Grad)} & 21.69 & 0.768 & 0.358 & 65.33 & 2.90 & 91.90 & 96.31 & 10.83 \\
        & \mbox{OctreeSplat (Point)} & 21.55 & 0.756 & 0.344 & 59.08 & 2.67 & 91.95 & 96.53 & 18.16 \\
        & \mbox{OctreeSplat (Finest)} & 21.83 & 0.775 & 0.307 & 64.89 & 2.87 & 91.96 & 96.35 & 23.98 \\
        \bottomrule
    \end{tabularx}
        \end{minipage}%
    }
\end{table*}

\subsubsection{Depth Supervision}
We ablate depth supervision to assess the contribution of RGB-D training data. First, we remove the ground-truth depth and normal losses, $\mathcal{L}_{D}$ and $\mathcal{L}_{N}$, denoted OctreeSplat (w/o data depth). We then additionally remove the VFM depth and normal alignment losses, $\mathcal{L}_{D}^{\mathrm{VFM}}$ and $\mathcal{L}_{N}^{\mathrm{VFM}}$, denoted OctreeSplat (w/o all depth losses). 
As shown in \autoref{tab:ablation_studies}(b) shows that removing depth supervision generally slightly improves RGB metrics and slightly degrades depth metrics. The model retains useful geometric accuracy without explicit depth losses, supporting its applicability to RGB-only training.

\subsubsection{Voxel Size}
We further ablate the coarsest voxel size over a range of 6–64 mm. As shown in \autoref{tab:ablation_studies}(c), increasing the voxel size substantially degrades RGB synthesis quality and reduces the number of Gaussian primitives, while its impact on depth synthesis is less pronounced. Balancing reconstruction quality and computational cost, and following prior work, we set the default coarsest voxel size to 8 mm.

\subsubsection{Octree Subdivision}

\begin{figure}[!htbp]
    \centering
    \includegraphics[width=\linewidth]{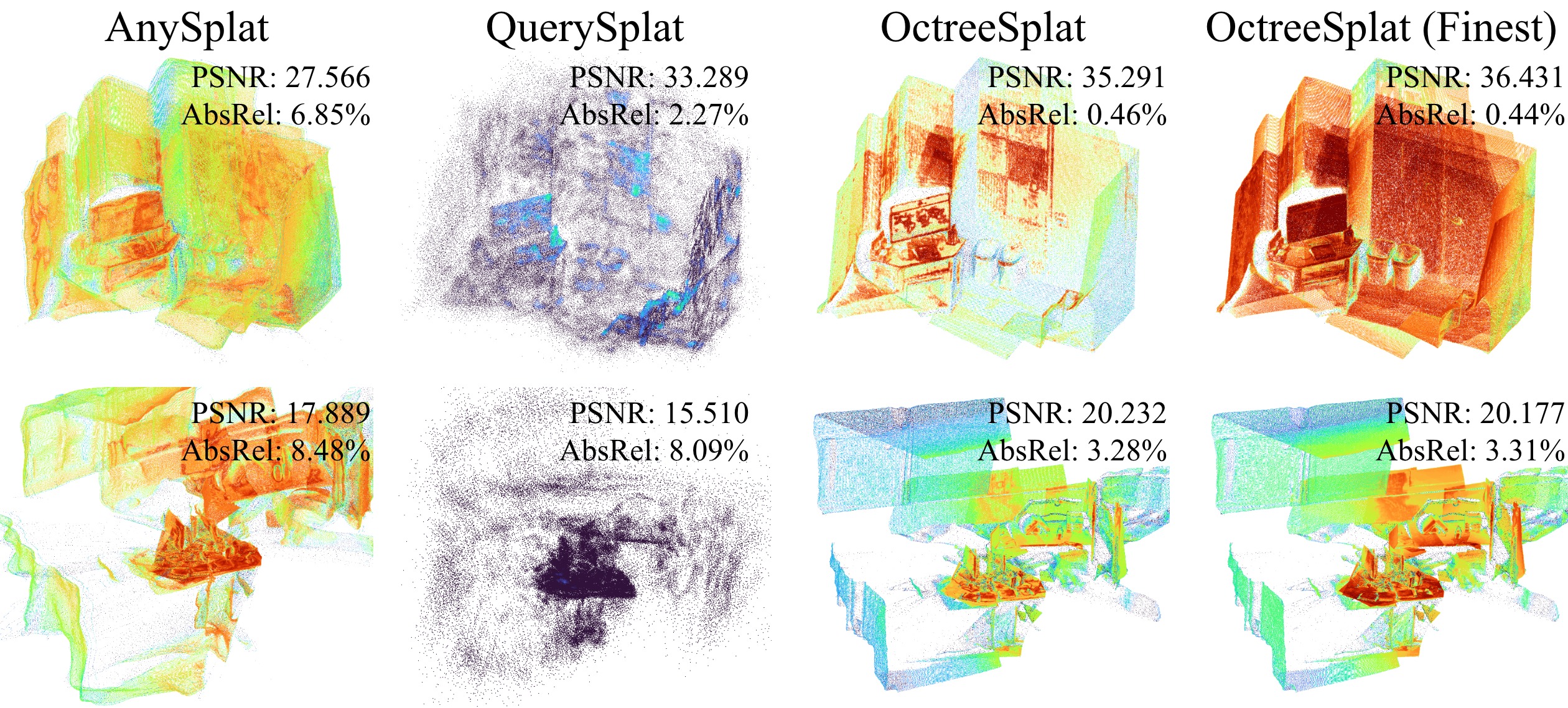}
    \caption{Spatial density of Gaussian primitives. Redder regions indicate higher primitive density.}
    \label{fig:point_density}
\end{figure}

We compare four primitive-allocation strategies to assess score-guided octree subdivision. OctreeSplat uses predicted densification scores. OctreeSplat (Grad) renders multiple views and computes gradients directly to guide subdivision. OctreeSplat (Point) uses the maximum multi-view sampling rate, while OctreeSplat (Finest) uses Gaussian primitives from the finest voxel level. All subdivision-based variants use two rounds, splitting 30\% of the voxels per round and discarding children without spatial points.

As shown in \autoref{tab:ablation_studies}(d), OctreeSplat (Finest) achieves the highest RGB and depth synthesis accuracy but uses the most Gaussian primitives. OctreeSplat (Point) exhibits a substantial performance drop without a notable reduction in primitive count. OctreeSplat and OctreeSplat (Grad) achieve comparable performance and primitive counts, while OctreeSplat avoids the multi-view rendering and gradient computation required by OctreeSplat (Grad). These results demonstrate the effectiveness of the densification score guided octree subdivision strategy. As shown in \autoref{fig:point_density}, OctreeSplat appropriately allocates more primitives to complex regions.

\FloatBarrier
\section{Conclusion and Future Work}
We presented OctreeSplat, a hierarchical octree-guided method for feed-forward 3DGS reconstruction.
Given images and optional camera parameters, OctreeSplat constructs hierarchical voxel features from multi-view features and geometric priors provided by a frozen VFM. A sparse-convolutional U-Net and separate heads jointly predict Gaussian parameters and inter-level densification scores. These scores guide octree subdivision and allocate primitives according to local scene complexity.
Training updates only 76 million parameters in the feature DPT, SPUNet, densification-score heads, and Gaussian prediction heads, and takes 35 GPU-hours.
Stochastic-solid volume rendering improves depth rendering and geometric supervision. Optional TTO optimizes LoRA modules inserted into the frozen backbone to align predicted and ground-truth camera parameters.
Experiments across multiple datasets demonstrate strong RGB synthesis and geometric accuracy, while octree-guided allocation produces more compact scene representations. When camera parameters are supplied, TTO improves pose accuracy and alignment between reconstructed and ground-truth surfaces.

Future work will investigate more flexible subdivision strategies and dynamic scene reconstruction. The current octree uses predefined levels and fixed subdivision ratios. Learning adaptive subdivision and merging from local geometric complexity may offer greater flexibility in balancing compactness and reconstruction quality.
The present method primarily targets static scenes and does not explicitly model object motion or nonrigid deformation.
Incorporating Gaussian motion models could extend feed-forward reconstruction to dynamic scenes while preserving geometric consistency across frames.

\section*{Acknowledgment}
The authors gratefully acknowledge the financial support from the Zhejiang
Provincial “Pioneer” and “Leading Goose + X” Science and Technology
Program (Grant No. 2026C02A1220), the Shanghai Science and Technology
Action Plan (Grant No. 21JM0010300), the Shanghai Aerospace Science
and Technology Innovation Fund (SAST) (Grant No. 2021-037), the Special
Fund for Technology Innovation Support Projects of Shanghai (Grant
Nos. 2021-cyxt-kj1, XTCX-KJ-2022-37, and HCXBCY-2023-046), and the
National Defense Basic Scientific Research Program of China (Grant No.
JCKY2021606B002).

\bibliographystyle{IEEEtran}
\bibliography{bib/sample}

@inproceedings{wang2025vggt,
  title={VGGT: Visual Geometry Grounded Transformer},
  author={Wang, Jianyuan and Chen, Minghao and Karaev, Nikita and Vedaldi, Andrea and Rupprecht, Christian and Novotny, David},
  booktitle={Proceedings of the IEEE/CVF Conference on Computer Vision and Pattern Recognition},
  year={2025}
}

@inproceedings{depthanything3,
 author = {Lin, Haotong and Chen, Sili and Liew, Jun Hao and Chen, Donny Y. and Li, Zhenyu and Zhao, Yang and Peng, Sida and Guo, Hengkai and Zhou, Xiaowei and Shi, Guang and Feng, Jiashi and Kang, Bingyi},
 booktitle = {International Conference on Learning Representations},
 editor = {C. Vondrick and B. Hariharan and C. Raffel and L. Pinto and D. Yang and A. Faust},
 pages = {141261--141285},
 title = {Depth Anything 3: Recovering the Visual Space from Any Views},
 volume = {2026},
 year = {2026}
}

@article{DPT,
	author    = {Ren\'{e} Ranftl and Alexey Bochkovskiy and Vladlen Koltun},
	title     = {Vision Transformers for Dense Prediction},
	journal   = {ArXiv preprint},
	year      = {2021},
}

@misc{oquab2023dinov2,
  title={DINOv2: Learning Robust Visual Features without Supervision},
  author={Oquab, Maxime and Darcet, Timothée and Moutakanni, Theo and Vo, Huy V. and Szafraniec, Marc and Khalidov, Vasil and Fernandez, Pierre and Haziza, Daniel and Massa, Francisco and El-Nouby, Alaaeldin and Howes, Russell and Huang, Po-Yao and Xu, Hu and Sharma, Vasu and Li, Shang-Wen and Galuba, Wojciech and Rabbat, Mike and Assran, Mido and Ballas, Nicolas and Synnaeve, Gabriel and Misra, Ishan and Jegou, Herve and Mairal, Julien and Labatut, Patrick and Joulin, Armand and Bojanowski, Piotr},
  journal={arXiv:2304.07193},
  year={2023}
}

@article{2023Objectsasvolumes,
  title={Objects as volumes: A stochastic geometry view of opaque solids},
  author={ Miller, Bailey  and  Chen, Hanyu  and  Lai, Alice  and  Gkioulekas, Ioannis },
  journal={IEEE},
  year={2023},
}

@article{zhang2026gggs,
  title={Geometry-Grounded Gaussian Splatting},
  author={Zhang, Baowen and Jiang, Chenxing and Li, Heng and Shen, Shaojie and Tan, Ping},
  journal={arXiv preprint arXiv:2601.17835},
  year={2026}
}

@Article{kerbl3DGS,
      author       = {Kerbl, Bernhard and Kopanas, Georgios and Leimk{\"u}hler, Thomas and Drettakis, George},
      title        = {3D Gaussian Splatting for Real-Time Radiance Field Rendering},
      journal      = {ACM Transactions on Graphics},
      number       = {4},
      volume       = {42},
      month        = {July},
      year         = {2023},
      url          = {https://repo-sam.inria.fr/fungraph/3d-gaussian-splatting/}
}

@inproceedings{Huang2DGS2024,
    title={2D Gaussian Splatting for Geometrically Accurate Radiance Fields},
    author={Huang, Binbin and Yu, Zehao and Chen, Anpei and Geiger, Andreas and Gao, Shenghua},
    publisher = {Association for Computing Machinery},
    booktitle = {SIGGRAPH 2024 Conference Papers},
    year      = {2024},
    doi       = {10.1145/3641519.3657428}
}

@inproceedings{dust3r_cvpr24,
      title={DUSt3R: Geometric 3D Vision Made Easy}, 
      author={Shuzhe Wang and Vincent Leroy and Yohann Cabon and Boris Chidlovskii and Jerome Revaud},
      booktitle = {CVPR},
      year = {2024}
}

@inproceedings{mildenhall2020nerf,
  title={NeRF: Representing Scenes as Neural Radiance Fields for View Synthesis},
  author={Ben Mildenhall and Pratul P. Srinivasan and Matthew Tancik and Jonathan T. Barron and Ravi Ramamoorthi and Ren Ng},
  year={2020},
  booktitle={ECCV},
}

@inproceedings{wang2025pi3,
  title={{$\pi^3$}: Permutation-Equivariant Visual Geometry Learning},
  author={Wang, Yifan and Zhou, Jianjun and Zhu, Haoyi and Chang, Wenzheng and Zhou, Yang and Li, Zizun and Chen, Junyi and Pang, Jiangmiao and Shen, Chunhua and He, Tong},
  booktitle={International Conference on Learning Representations},
  volume={2026},
  pages={10481--10497},
  year={2026}
}

@inproceedings{barron2022mipnerf360,
  title={Mip-nerf 360: Unbounded anti-aliased neural radiance fields},
  author={Barron, Jonathan T and Mildenhall, Ben and Verbin, Dor and Srinivasan, Pratul P and Hedman, Peter},
  booktitle={Proceedings of the IEEE/CVF conference on computer vision and pattern recognition},
  pages={5470--5479},
  year={2022}
}

@inproceedings{colmapsfm,
    author={Sch\"{o}nberger, Johannes Lutz and Frahm, Jan-Michael},
    title={Structure-from-Motion Revisited},
    booktitle={Conference on Computer Vision and Pattern Recognition (CVPR)},
    year={2016},
}

@inproceedings{li2026tokensplat,
  title={Tokensplat: Token-aligned 3d gaussian splatting for feed-forward pose-free reconstruction},
  author={Li, Yihui and Lv, Chengxin and Tang, Zichen and Yang, Hongyu and Huang, Di},
  booktitle={Proceedings of the IEEE/CVF Conference on Computer Vision and Pattern Recognition},
  pages={40886--40895},
  year={2026}
}

@article{jiang2025anysplat,
  title={Anysplat: Feed-forward 3d gaussian splatting from unconstrained views},
  author={Jiang, Lihan and Mao, Yucheng and Xu, Linning and Lu, Tao and Ren, Kerui and Jin, Yichen and Xu, Xudong and Yu, Mulin and Pang, Jiangmiao and Zhao, Feng and others},
  journal={ACM Transactions on Graphics (TOG)},
  volume={44},
  number={6},
  pages={1--16},
  year={2025},
  publisher={ACM New York, NY, USA}
}

@inproceedings{ye2025yonosplatneedmodelfeedforward,
  title     = {YoNoSplat: You Only Need One Model for Feedforward 3D Gaussian Splatting},
  author    = {Ye, Botao and Chen, Boqi and Xu, Haofei and Barath, Daniel and Pollefeys, Marc},
  booktitle = {International Conference on Learning Representations (ICLR)},
  year      = {2026}
}

@inproceedings{xu2024depthsplat,
      title   = {DepthSplat: Connecting Gaussian Splatting and Depth},
      author  = {Xu, Haofei and Peng, Songyou and Wang, Fangjinhua and Blum, Hermann and Barath, Daniel and Geiger, Andreas and Pollefeys, Marc},
      booktitle={CVPR},
      year={2025}
    }

@inproceedings{chen2024mvsplat,
  title={Mvsplat: Efficient 3d gaussian splatting from sparse multi-view images},
  author={Chen, Yuedong and Xu, Haofei and Zheng, Chuanxia and Zhuang, Bohan and Pollefeys, Marc and Geiger, Andreas and Cham, Tat-Jen and Cai, Jianfei},
  booktitle={European conference on computer vision},
  pages={370--386},
  year={2024},
  organization={Springer}
}

@inproceedings{wang2025volsplat,
  title={Volsplat: Rethinking feed-forward 3d gaussian splatting with voxel-aligned prediction},
  author={Wang, Weijie and Chen, Yeqing and Zhang, Zeyu and Liu, Hengyu and Wang, Haoxiao and Feng, Zhiyuan and Qin, Wenkang and Chen, Feng and Bian, Jia-Wang and Zhu, Zheng and others},
  booktitle={European Conference on Computer Vision},
  pages={1--19},
  year={2026},
  organization={Springer}
}

@inproceedings{infinigen2023infinite,
  title={Infinite Photorealistic Worlds Using Procedural Generation},
  author={Raistrick, Alexander and Lipson, Lahav and Ma, Zeyu and Mei, Lingjie and Wang, Mingzhe and Zuo, Yiming and Kayan, Karhan and Wen, Hongyu and Han, Beining and Wang, Yihan and Newell, Alejandro and Law, Hei and Goyal, Ankit and Yang, Kaiyu and Deng, Jia},
  booktitle={Proceedings of the IEEE/CVF Conference on Computer Vision and Pattern Recognition},
  pages={12630--12641},
  year={2023}
}

@article{replica19arxiv,
  title =   {The {R}eplica Dataset: A Digital Replica of Indoor Spaces},
  author =  {Julian Straub and Thomas Whelan and Lingni Ma and Yufan Chen and Erik Wijmans and Simon Green and Jakob J. Engel and Raul Mur-Artal and Carl Ren and Shobhit Verma and Anton Clarkson and Mingfei Yan and Brian Budge and Yajie Yan and Xiaqing Pan and June Yon and Yuyang Zou and Kimberly Leon and Nigel Carter and Jesus Briales and  Tyler Gillingham and  Elias Mueggler and Luis Pesqueira and Manolis Savva and Dhruv Batra and Hauke M. Strasdat and Renzo De Nardi and Michael Goesele and Steven Lovegrove and Richard Newcombe },
  journal = {arXiv preprint arXiv:1906.05797},
  year =    {2019}
}

@article{dtu,
  title={Large-Scale Data for Multiple-View Stereopsis},
  author={Aanaes and HenrikJensen and Rasmus RamsbolVogiatzis and GeorgeTola and EnginDahl and Anders Bjorholm},
  journal={International Journal of Computer Vision},
  volume={120},
  number={2},
  year={2016},
}

@article{TAT,
    author    = {Arno Knapitsch and Jaesik Park and Qian-Yi Zhou and Vladlen Koltun},
    title     = {Tanks and Temples: Benchmarking Large-Scale Scene Reconstruction},
    journal   = {ACM Transactions on Graphics},
    volume    = {36},
    number    = {4},
    year      = {2017},
}

@article{SSIM,
  title={Image quality assessment: from error visibility to structural similarity},
  author={ Wang, Zhou  and  Bovik, Alan Conrad  and  Sheikh, Hamid Rahim  and  Simoncelli, Eero P. },
  journal={IEEE Trans Image Process},
  volume={13},
  number={4},
  year={2004},
}

@article{LPIPS,
  title={The Unreasonable Effectiveness of Deep Features as a Perceptual Metric},
  author={ Zhang, Richard  and  Isola, Phillip  and  Efros, Alexei A  and  Shechtman, Eli  and  Wang, Oliver },
  journal={IEEE},
  year={2018},
}

@misc{wang2026vggtomega,
      title={VGGT-$\Omega$}, 
      author={Jianyuan Wang and Minghao Chen and Shangzhan Zhang and Nikita Karaev and Johannes Schönberger and Patrick Labatut and Piotr Bojanowski and David Novotny and Andrea Vedaldi and Christian Rupprecht},
      year={2026},
      eprint={2605.15195}
}

@misc{spconv2022,
    title={Spconv: Spatially Sparse Convolution Library},
    author={Spconv Contributors},
    howpublished = {\url{https://github.com/traveller59/spconv}},
    year={2022}
}

@article{umeyama1991least,
  title={Least-Squares Estimation of Transformation Parameters Between Two Point Patterns},
  author={Umeyama, Shinji},
  journal={IEEE Transactions on Pattern Analysis and Machine Intelligence},
  volume={13},
  number={4},
  pages={376--380},
  year={1991},
  publisher={IEEE}
}

@misc{kim2026f4splat,
  title         = {F4Splat: Feed-Forward Predictive Densification for Feed-Forward 3D Gaussian Splatting},
  author        = {Kim, Injae and Kim, Chaehyeon and Bae, Minseong and Joo, Minseok and Kim, Hyunwoo J.},
  year          = {2026},
  eprint        = {2603.21304},
  archivePrefix = {arXiv},
  primaryClass  = {cs.CV},
  url           = {https://arxiv.org/abs/2603.21304}
}

@article{li2026querysplat,
  title={QuerySplat: Decoupling Geometry and Appearance Representations in 3DGS Prediction},
  author={Li, Yinglong and Shen, Donghui and Zhang, Xiaoyu and Ye, Zhichao and Wu, Hongyu and Hao, Aimin and Zhang, Guofeng and Liu, Haomin},
  journal={arXiv preprint arXiv:2608.01186},
  year={2026},
  url={https://arxiv.org/abs/2608.01186}
}

@inproceedings{svraster,
  author       = {Cheng Sun and
                  Jaesung Choe and
                  Charles Loop and
                  Wei{-}Chiu Ma and
                  Yu{-}Chiang Frank Wang},
  title        = {Sparse Voxels Rasterization: Real-time High-fidelity Radiance Field
                  Rendering},
  booktitle    = {{CVPR}},
  year         = {2025},
}

@inproceedings{yeshwanth2023scannet++,
  title={Scannet++: A high-fidelity dataset of 3d indoor scenes},
  author={Yeshwanth, Chandan and Liu, Yueh-Cheng and Nie{\ss}ner, Matthias and Dai, Angela},
  booktitle={2023 IEEE/CVF International Conference on Computer Vision (ICCV)},
  pages={12--22},
  year={2023},
  organization={IEEE}
}

@article{baruch2021arkitscenes,
  title={Arkitscenes: A diverse real-world dataset for 3d indoor scene understanding using mobile rgb-d data},
  author={Baruch, Gilad and Chen, Zhuoyuan and Dehghan, Afshin and Dimry, Tal and Feigin, Yuri and Fu, Peter and Gebauer, Thomas and Joffe, Brandon and Kurz, Daniel and Schwartz, Arik and others},
  journal={arXiv preprint arXiv:2111.08897},
  year={2021}
}

@inproceedings{ling2024dl3dv,
  title={Dl3dv-10k: A large-scale scene dataset for deep learning-based 3d vision},
  author={Ling, Lu and Sheng, Yichen and Tu, Zhi and Zhao, Wentian and Xin, Cheng and Wan, Kun and Yu, Lantao and Guo, Qianyu and Yu, Zixun and Lu, Yawen and others},
  booktitle={2024 IEEE/CVF Conference on Computer Vision and Pattern Recognition (CVPR)},
  pages={22160--22169},
  year={2024},
  organization={IEEE}
}

@inproceedings{ye2024absgs,
  title={Absgs: Recovering fine details in 3d gaussian splatting},
  author={Ye, Zongxin and Li, Wenyu and Liu, Sidun and Qiao, Peng and Dou, Yong},
  booktitle={Proceedings of the 32nd ACM international conference on multimedia},
  pages={1053--1061},
  year={2024}
}

@inproceedings{tum2012benchmark,
  title={A benchmark for the evaluation of RGB-D SLAM systems},
  author={Sturm, J{\"u}rgen and Engelhard, Nikolas and Endres, Felix and Burgard, Wolfram and Cremers, Daniel},
  booktitle={2012 IEEE/RSJ international conference on intelligent robots and systems},
  pages={573--580},
  year={2012},
  organization={IEEE}
}

@inproceedings{charatan2024pixelsplat,
  title={pixelsplat: 3d gaussian splats from image pairs for scalable generalizable 3d reconstruction},
  author={Charatan, David and Li, Sizhe Lester and Tagliasacchi, Andrea and Sitzmann, Vincent},
  booktitle={2024 IEEE/CVF Conference on Computer Vision and Pattern Recognition (CVPR)},
  pages={19457--19467},
  year={2024},
  organization={IEEE}
}

@book{BA,
  title={Multiple view geometry in computer vision},
  author={Hartley, Richard and Zisserman, Andrew and others},
  volume={2},
  number={3},
  year={2003},
  publisher={Cambridge university press, Cambridge}
}

@article{seitz1999photorealistic,
  title={Photorealistic scene reconstruction by voxel coloring},
  author={Seitz, Steven M and Dyer, Charles R},
  journal={International journal of computer vision},
  volume={35},
  number={2},
  pages={151--173},
  year={1999},
  publisher={Springer}
}

@inproceedings{curless1996volumetric,
  title={A volumetric method for building complex models from range images},
  author={Curless, Brian and Levoy, Marc},
  booktitle={Proceedings of the 23rd annual conference on Computer graphics and interactive techniques},
  pages={303--312},
  year={1996}
}

@inproceedings{labatut2009robust,
  title={Robust and efficient surface reconstruction from range data},
  author={Labatut, Patrick and Pons, J-P and Keriven, Renaud},
  booktitle={Computer graphics forum},
  volume={28},
  number={8},
  pages={2275--2290},
  year={2009},
  organization={Wiley Online Library}
}

@article{mueller2022instantngp,
    author = {Thomas M\"uller and Alex Evans and Christoph Schied and Alexander Keller},
    title = {Instant Neural Graphics Primitives with a Multiresolution Hash Encoding},
    journal = {ACM Trans. Graph.},
    issue_date = {July 2022},
    volume = {41},
    number = {4},
    month = jul,
    year = {2022},
    pages = {102:1--102:15},
    articleno = {102},
    numpages = {15},
    url = {https://doi.org/10.1145/3528223.3530127},
    doi = {10.1145/3528223.3530127},
    publisher = {ACM},
    address = {New York, NY, USA},
}

@inproceedings{yu2022plenoxels,
      title={Plenoxels: Radiance Fields without Neural Networks}, 
      author={Sara Fridovich-Keil and Alex Yu and Matthew Tancik and Qinhong Chen and Benjamin Recht and Angjoo Kanazawa},
      year={2022},
      booktitle={CVPR},
}

@misc{barron2021mipnerf,
      title={Mip-NeRF: A Multiscale Representation for Anti-Aliasing Neural Radiance Fields},
      author={Jonathan T. Barron and Ben Mildenhall and Matthew Tancik and Peter Hedman and Ricardo Martin-Brualla and Pratul P. Srinivasan},
      year={2021},
      eprint={2103.13415},
      archivePrefix={arXiv},
      primaryClass={cs.CV}
}

@article{wang2021neus,
  title={NeuS: Learning Neural Implicit Surfaces by Volume Rendering for Multi-view Reconstruction},
  author={Wang, Peng and Liu, Lingjie and Liu, Yuan and Theobalt, Christian and Komura, Taku and Wang, Wenping},
  journal={arXiv preprint arXiv:2106.10689},
  year={2021}
}

@inproceedings{neus2,
    title={NeuS2: Fast Learning of Neural Implicit Surfaces for Multi-view Reconstruction}, 
    author={Wang, Yiming and Han, Qin and Habermann, Marc and Daniilidis, Kostas and Theobalt, Christian and Liu, Lingjie},
    year={2023},
    booktitle={Proceedings of the IEEE/CVF International Conference on Computer Vision (ICCV)}
}

@inproceedings{kim2024color,
  title={Color-cued efficient densification method for 3d gaussian splatting},
  author={Kim, Sieun and Lee, Kyungjin and Lee, Youngki},
  booktitle={2024 IEEE/CVF Conference on Computer Vision and Pattern Recognition Workshops (CVPRW)},
  pages={775--783},
  year={2024},
  organization={IEEE}
}

@inproceedings{lyu2025resgs,
  title={Resgs: Residual densification of 3d gaussian for efficient detail recovery},
  author={Lyu, Yanzhe and Cheng, Kai and Kang, Xin and Chen, Xuejin},
  booktitle={2025 IEEE/CVF International Conference on Computer Vision (ICCV)},
  pages={28093--28102},
  year={2025},
  organization={IEEE}
}

@inproceedings{chen2025dashgaussian,
  title={Dashgaussian: Optimizing 3d gaussian splatting in 200 seconds},
  author={Chen, Youyu and Jiang, Junjun and Jiang, Kui and Tang, Xiao and Li, Zhihao and Liu, Xianming and Nie, Yinyu},
  booktitle={2025 IEEE/CVF Conference on Computer Vision and Pattern Recognition (CVPR)},
  pages={11146--11155},
  year={2025},
  organization={IEEE}
}

@inproceedings{mallick2024taming3dgs,
  title={Taming 3dgs: High-quality radiance fields with limited resources},
  author={Mallick, Saswat Subhajyoti and Goel, Rahul and Kerbl, Bernhard and Steinberger, Markus and Carrasco, Francisco Vicente and De La Torre, Fernando},
  booktitle={SIGGRAPH Asia 2024 Conference Papers},
  pages={1--11},
  year={2024}
}

@article{meuleman2025fly,
  title={On-the-fly reconstruction for large-scale novel view synthesis from unposed images},
  author={Meuleman, Andreas and Shah, Ishaan and Lanvin, Alexandre and Kerbl, Bernhard and Drettakis, George},
  journal={ACM Transactions on Graphics (TOG)},
  volume={44},
  number={4},
  pages={1--14},
  year={2025},
  publisher={{ACM, New York, NY, USA}}
}

@inproceedings{ren2026fastgs,
  title={Fastgs: Training 3d gaussian splatting in 100 seconds},
  author={Ren, Shiwei and Wen, Tianci and Fang, Yongchun and Lu, Biao},
  booktitle={Proceedings of the IEEE/CVF Conference on Computer Vision and Pattern Recognition},
  pages={26094--26103},
  year={2026}
}

@inproceedings{cabon2025must3r,
  title={Must3r: Multi-view network for stereo 3d reconstruction},
  author={Cabon, Yohann and Stoffl, Lucas and Antsfeld, Leonid and Csurka, Gabriela and Chidlovskii, Boris and Revaud, Jerome and Leroy, Vincent},
  booktitle={2025 IEEE/CVF Conference on Computer Vision and Pattern Recognition (CVPR)},
  pages={1050--1060},
  year={2025},
  organization={IEEE}
}

@article{liu2025vggtx,
  title={Vggt-x: When vggt meets dense novel view synthesis},
  author={Liu, Yang and Luo, Chuanchen and Tang, Zimo and Peng, Junran and Zhang, Zhaoxiang},
  journal={arXiv preprint arXiv:2509.25191},
  year={2025}
}

@article{smart2024splatt3r,
      journal={arXiv preprint arXiv:2408.13912},
      title={Splatt3R: Zero-shot Gaussian Splatting from Uncalibrated Image Pairs}, 
      author={Brandon Smart and Chuanxia Zheng and Iro Laina and Victor Adrian Prisacariu},
      year={2024},
      eprint={2408.13912},
      archivePrefix={arXiv},
      primaryClass={cs.CV},
      url={https://arxiv.org/abs/2408.13912}, 
}

@article{guedon2023sugar,
  title={SuGaR: Surface-Aligned Gaussian Splatting for Efficient 3D Mesh Reconstruction and High-Quality Mesh Rendering},
  author={Gu{\'e}don, Antoine and Lepetit, Vincent},
  journal={CVPR},
  year={2024}
}

@article{chen2024pgsr,
  title={PGSR: Planar-based Gaussian Splatting for Efficient and High-Fidelity Surface Reconstruction},
  author={Chen, Danpeng and Li, Hai and Ye, Weicai and Wang, Yifan and Xie, Weijian and Zhai, Shangjin and Wang, Nan and Liu, Haomin and Bao, Hujun and Zhang, Guofeng},
  journal={arXiv preprint arXiv:2406.06521},
  year={2024}
}

@article{Yu2024GOF,
  author    = {Yu, Zehao and Sattler, Torsten and Geiger, Andreas},
  title     = {Gaussian Opacity Fields: Efficient Adaptive Surface Reconstruction in Unbounded Scenes},
  journal   = {ACM Transactions on Graphics},
  year      = {2024},
}

@misc{gw,
      title={From Blobs to Spokes: High-Fidelity Surface Reconstruction via Oriented Gaussians}, 
      author={Diego Gomez and Antoine Guédon and Nissim Maruani and Bingchen Gong and Maks Ovsjanikov},
      year={2026},
      eprint={2604.07337},
      archivePrefix={arXiv},
      primaryClass={cs.CV},
      url={https://arxiv.org/abs/2604.07337}, 
}
\end{document}